\documentclass[conference,letterpaper]{IEEEtran}

\usepackage{times,mathptm}     
\usepackage{epsfig,cite}
\usepackage[T1]{fontenc}

\newcommand\be{\vspace*{-1pt}\begin{equation}}
\newcommand\ee{\end{equation}\vspace*{-1pt}}
\newcommand\bea{\vspace*{1pt}\begin{eqnarray}}
\newcommand\eea{\end{eqnarray}\vspace*{1pt}}
\newcommand{\barr}{\begin{array}}
\newcommand{\earr}{\end{array}}

\def\bA{{\mathbf A}}
\def\bB{{\mathbf B}}
\def\bX{{\mathbf X}}  \def\bx{{\mathbf x}}
\def\bO{{\mathbf O}}

\begin{document}

\title{\Large\bf Coloring black boxes: visualization of neural network decisions.}

\author{W{\l}odzis{\l}aw Duch\\
School of Computer Engineering, Nanyang Technological University, Singapore\\
\& Dept. of Informatics, Nicolaus Copernicus University, Toru\'n, Poland \\
\ \ \ \ http://www.is.umk.pl/~duch/
}

\maketitle              

\thispagestyle{empty} \pagestyle{empty}

\begin{abstract}
Neural networks are commonly regarded as black boxes performing incomprehensible functions.
For classification problems networks provide maps from high dimensional feature space to $K$-dimensional image space. Images of training vector are projected on polygon vertices, providing visualization of network function. Such visualization may show the dynamics of learning, allow for comparison of different networks, display training vectors around which potential problems may arise, show differences due to regularization and optimization procedures, investigate stability of network classification under perturbation of original vectors, and place new data sample in relation to training data, allowing for estimation of confidence in classification of a given sample. An illustrative example for the three-class Wine data and five-class Satimage data is described. The visualization method proposed here is applicable to any black box system that provides continuous outputs.
\end{abstract}


\section{Introduction}

In common opinion neural networks are black boxes that should not be used for safety-critical applications. Some understanding of network decisions may be found if the network is converted to logical rules \cite{duchtnn}. This understanding always comes at a price. If network function is approximated decision borders provided by neural networks are severely distorted, since feature space has to be partitioned into hypercuboids (for crisp logical rules) or ellipsoids (for typical triangular or Gaussian fuzzy membership functions). An alternative is to convert the neural network itself to a simplified structure performing logical functions. Since neural networks are universal approximators, and regularization leads to low-complexity models that perform quite well providing estimation of posterior probabilities, approximation by logical rules always distorts the mapping found by the network. Although for some data classification accuracy obtained with optimized logical rules is higher than the accuracy obtained by neural networks, it seems to be an artifact of quantization of outputs (for example, forcing the patient into "healthy" or "sick" categories)  \cite{duchtnn}.

What information do we get from a typical neural network? Estimation of the overall classification accuracy, mean square error (MSE), and sometimes estimation of the classification probability. The quality of two networks is compared only by looking at their accuracy, or at best at the Receiver Operator Characteristics (ROC) curves \cite{ROC}. All such measures are global; they do not distinguish between easy and difficult cases. Overall classification accuracy is not a good estimator of the accuracy for the particular problem at hand, since all errors may be confined to a distant and localized region of the feature space.
Multilayer Perceptron (MLP) networks provide outputs close to 0 and 1, making them overconfident in their predictions. There is a big difference between networks that make 10 errors, each time predicting wrong answer with probability close to 1, and networks that make the same wrong answers but with probability only slightly higher than that for the correct answer. Regularization may improve generalization \cite{Bishop} but since stochastic learning algorithms create networks with identical accuracy, but quite different weights and biases, which network should finally be choosen?
Is the network hidding some strange behavior that may lead to completely wrong results for new data?
Visualization of mappings performed by neural networks will certainly widen their range of applicability.


Since feature spaces are highly dimensional faithful presentation of the mapping learned by neural network is not possible. An interesting information is contained in perceived similarities of the training data samples. For classification problems with $K$ categories these similarities may be displayed as a scatterogram in $K$-dimensional space. In the next section a linear projection method is introduced, projecting the network outputs into $K$ vertices of a polygon. Section three presents a detailed case study using an MLP and RBF networks for the 3-class Wine dataset, and some examples for 5-class Satimage dataset. In the last section discussion and some remarks on the usefulness and further development of such visualization methods are given. Since the use of color makes it much easier to understand the figures the reader is advised to view the PDF version of the paper \cite{pdfversion}.

\section{Projection of network outputs.}

Assume that in $K$-class problem for each training vector $\bX$ neural network outputs $o_i(\bX)\in[0,1], i=1\dots K$ are given. They may come either from a single network, or $K$ networks with single output that specialize in discrimination of vectors from a single class.
The target output in a typical classification problem has $K-1$ zero outputs, and one $o_j(\bX)=1$ output that corresponds to the class $C_j$ the input vector $\bX$ belongs to. This requirement is in many cases artificial. The output classes may form continuum, rather then a small set of integer numbers, leading to a fuzzy ``degree of membership'' replacing crisp labeling. The outputs $o_i(\bX)$ may be treated as an estimation of this degree of membership, and in some caes as an estimation of similarity of  the vector $\bX$ to other vectors of the same class. In some network realizations the outputs are estimations of posterior probabilities $p(C_i|\bX;M)$, given the network $M$ and the vector $\bX$. Since probabilities sum to 1 the number of independent outputs is reduced to $K-1$. Networks outputs are $K$-dimensional images of inputs, created by the non-linear function that the network has learned. For vectors of different classes images created by neural networks that do not make any errors are separable clusters, otherwise these clusters will overlap.

Visualization of network decisions is possible in $K$-dimensional space, presenting images of all training vectors. For $K$=2, if the network outputs are independent (i.e. they do not sum to 1) the desired answers fall into $(1,0)$ and $(0,1)$ corners of a square in $(o_1,o_2)$ coordinates. Images of vectors that belong to the overlapping regions may be close to $(1,1)$ vertex, while vectors that are not recognized are close to $(0,0)$ vertex. Vectors $\bX$ that are far from decision borders and are classified correctly have scatterogram images $O(\bX)$ clustering around $(1,0)$ and $(0,1)$ corners. Images of vectors that are close to the decision borders fall closer to the middle of the square. Vectors from different classes are distinguished using different markers. Comparing such scatterograms for different networks will immediately show significant differences despite similar accuracies.
The position of the image of a new vector $\bX$ in relation to the images of training vectors shown in scatterogram allows for evaluation of the reliability of its classification.

\begin{figure}
\centering
\epsfig{figure=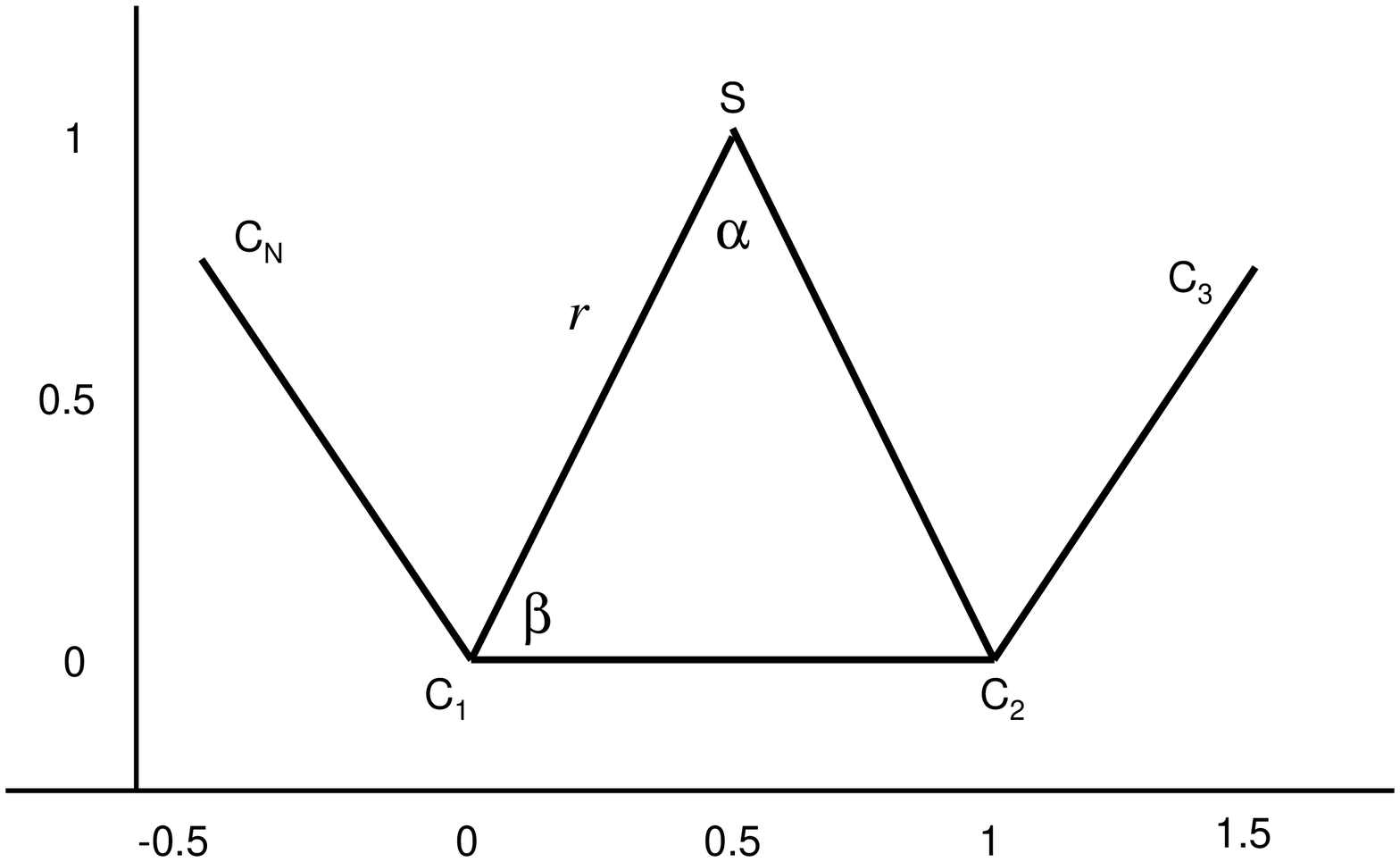,width=6.5cm}
\caption{Polygon used for projection of $K$-dimensional data.}
\label{fig:poly}
\end{figure}

Similar representation is possible for $K$=3, but for larger number of classes some projection on two or three dimensions is needed. Although all linear projections loose some information and more sophisticated projections could be devised, simple  approach presented below is already quite useful. The hypercube corners that correspond to binary labels (from $(1,0,..,0)$ to $(0,0,..,1)$) will correspond to $K$ corners of regular polygon in two dimensions. Coordinates of this polygon, with $(0,0)$ vertex corresponding to $(1,0,..,0)$ point, and $(0,1)$ vertex corresponding to $(0,1,..,0)$ point, are calculated from
(see Fig. \ref{fig:poly}):

\bea \label{eq:polygon}
\phi &=&-\frac{\pi}{2}-\frac{\pi}{K}, r=\frac{1}{2\cos(\frac{\pi}{2}-\frac{\pi}{K})} \nonumber \\
x_j &=& \frac{1}{2} +r\cos(\phi +\frac{2\pi j}{K}); \\
y_j &=& \frac{1}{2}\tan(\frac{\pi}{2}-\frac{\pi}{K}) +r\sin(\phi +\frac{2\pi j}{K}), j=0\dots K-1 \nonumber \\ \nonumber
\eea

The transformation $\bx=\bA \bO +\bB$ may be found by setting up $2K+2$ linear equations: $2K$ equations for projections of $(1,0,..,0)$ to $(0,0,..,1)$ unit vectors on $(x_j,y_j)$ polygon vertices, and two equations for projection of $(1,1,...,1)$ point on the polygon center $S$, with coordinates $(x_c,y_c)=\left(\frac{1}{2}, \frac{1}{2}\tan(\frac{\pi}{2}-\frac{\pi}{K}) \right)$.

\begin{figure}
\centering
\epsfig{figure=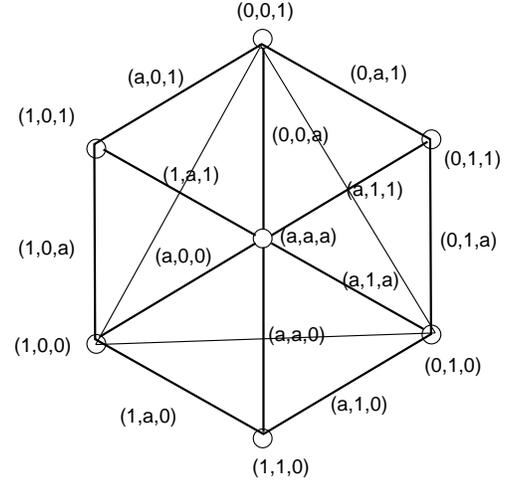,width=8.5cm}
\caption{Characteristic points and lines in the image space used for projection of $3$-dimensional data.}
\label{fig:poly3}
\end{figure}

This projection has several interesting features. For $K$=3 the center of the triangle corresponds to all $(a,a,a)$ points (where $a$ is arbitrary number) in 3 dimensions. Cases where all three outputs are 1 fall there, as well as cases where all three outputs are 0 (see Fig.\ref{fig:poly3}). Since all outputs are assumed to lie in the unit interval $[0,1]$, all points will lie within hexagon, with corners corresponding to binary $(o_1,o_2,o_3)$ values. The opposite corners of the hexagon have inverted bits, $\bar o_j=1-o_j$. Points corresponding to vectors that are weakly exciting $o_1$ output approach the center along the $(a,0,0)$ line, while points in the overlapping region of class two and three approach the center along the $(a,1,1)$ line.

\section{Case study: Wine data}

Chemical analysis of wines grown in the same region in Italy, but derived from three different cultivars, should be sufficient to recognize the source of the wine. The analysis determined 13 quantities, including alcohol content, hue, color intensity, and content 9 chemical compounds. The data is stored in UC Irvine repository of machine learning problems \cite{UCI}, where more details about it may be found.
The number of data samples from Classes 1, 2, and 3 is 59, 71 and 48, respectively, so the data is rather small. It is possible to separate the classes perfectly using an MLP network with just 2 hidden neurons. The 3 classes are designated by +, o and x markers.

The NETLAB neural network package \cite{Netlab} written in Matlab has been used in the experiments described below.  All MLP networks are trained with the scaled conjugate gradient procedure, with a single hidden layer network. These networks are used to map 13-dimensional vectors into 3-dimensions and then project the result to 2-dimensions using the method introduced in the previous section.
Using scatterograms of the training data created this way the following issues are addressed:

\begin{enumerate}
\item The dynamics of the neural learning.
\item Under and over-fitting effects.
\item Regularization effects.
\item Differences between networks of the same accuracy.
\end{enumerate}

\subsection{The dynamics of the neural learning}

\begin{figure}
\centering
\epsfig{figure=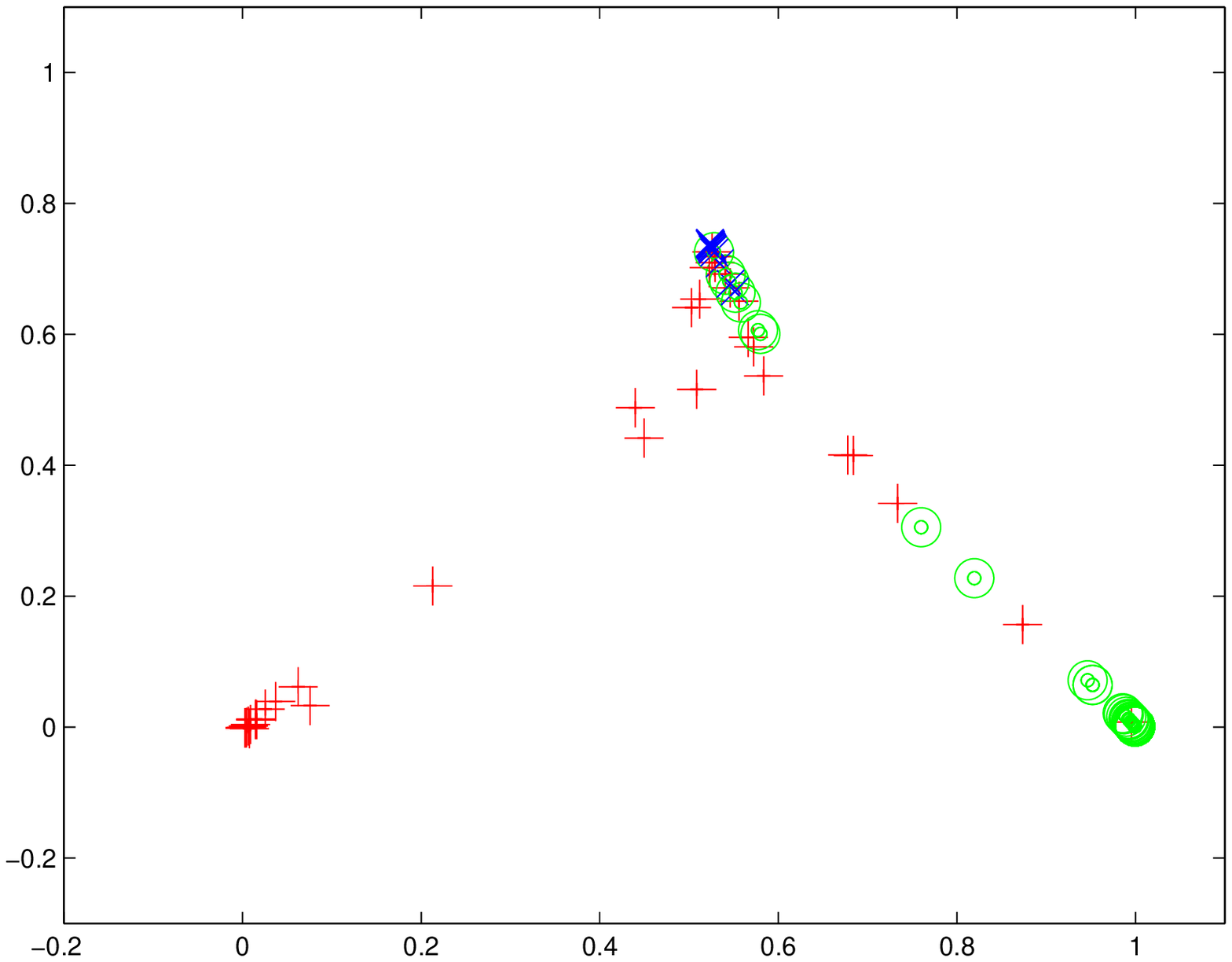,width=4.3cm}
\epsfig{figure=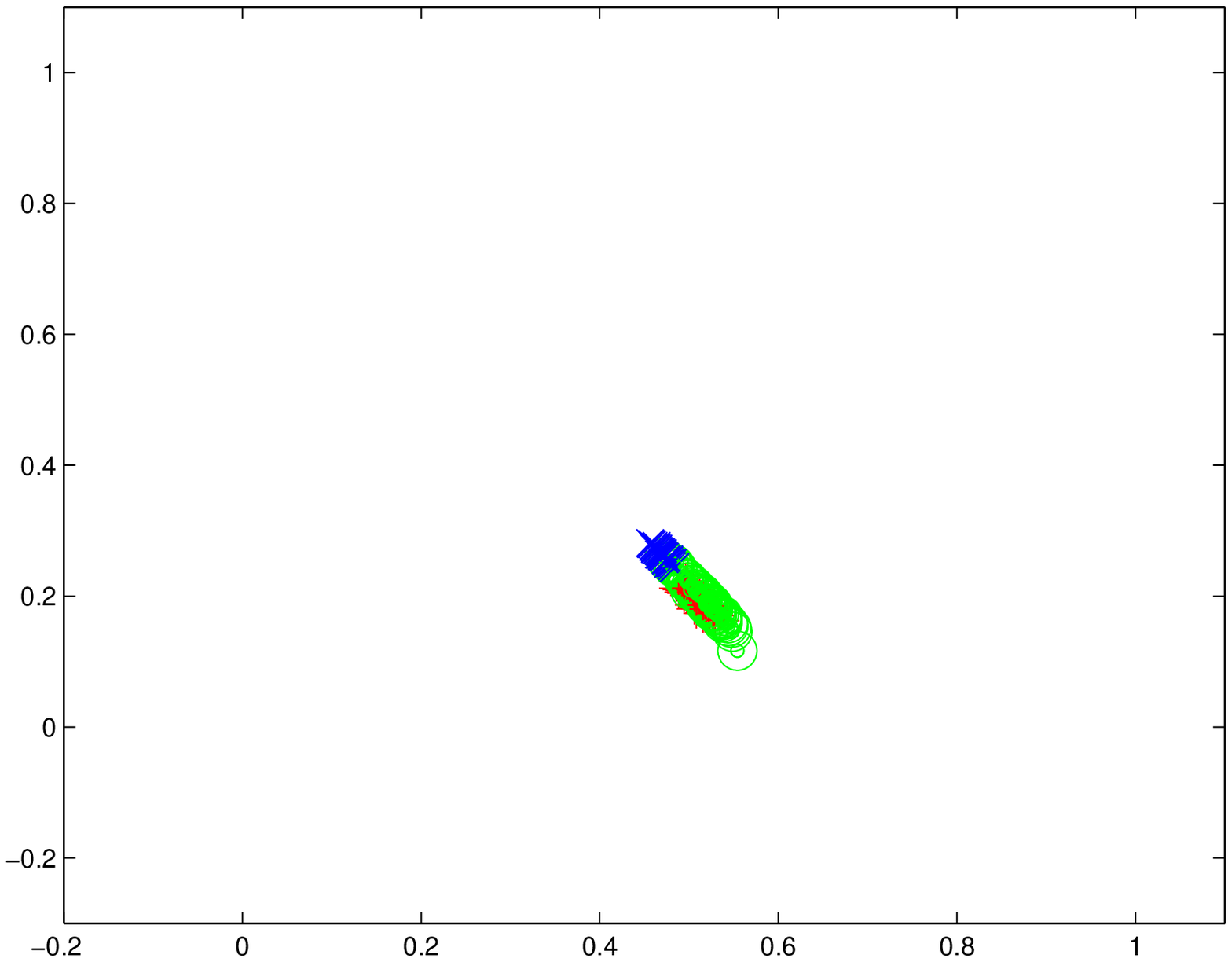,width=4.3cm}
\epsfig{figure=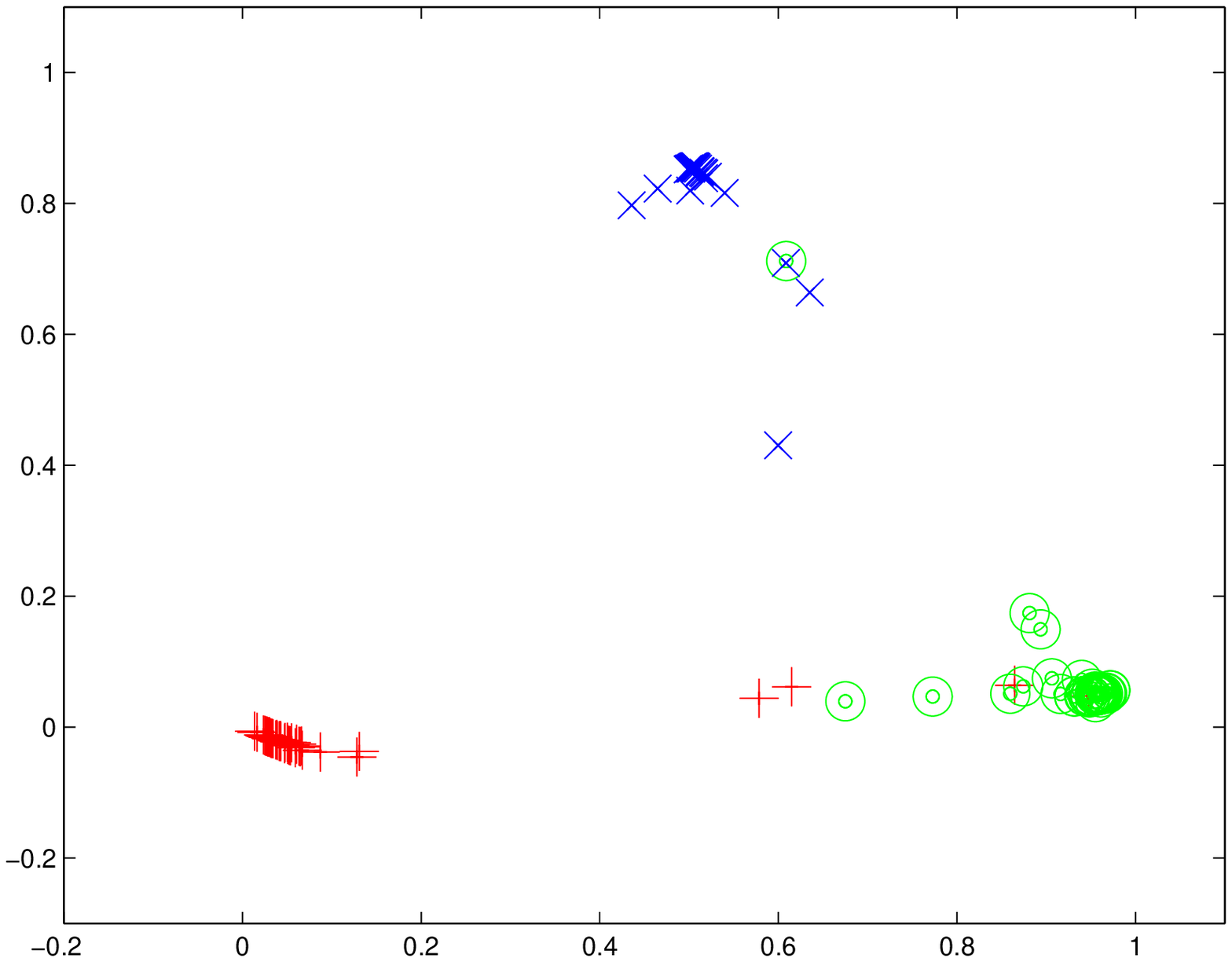,width=4.3cm}
\epsfig{figure=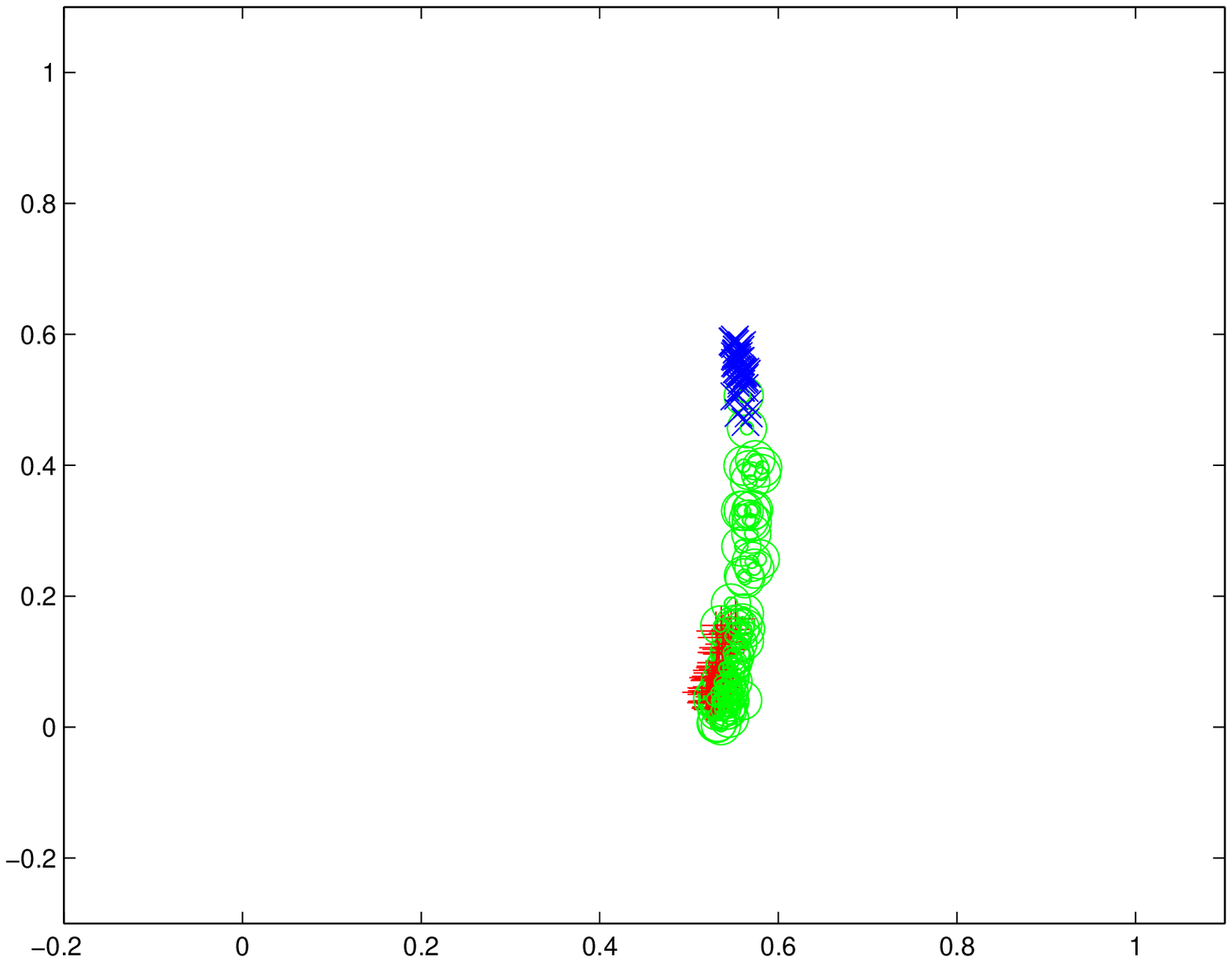,width=4.3cm}
\caption{Convergence of a network with 3 hidden neurons: top row - two solutions (27 and 107 errors) after 5 iterations, bottom row - two solutions (5 and 71 errors) after 10 iterations.}
\label{fig:conv1}
\end{figure}

Three hidden neurons have been used in numerical experiments here. Since the network is initialized with small values of weights and biases after the first training epoch all output values are concentrated around 0.5. The first series of pictures (Fig. \ref{fig:conv1}) shows the network performance after 5 and 10 iterations. Since each time the network is trained different solution is obtained two extreme cases were selected from 20 trials, the best network (lowest number of errors, on the left) and the worst network (largest number of errors, on the right). The vectors that are still not correctly handled are easily identified. In the lower left corner (0,0) most of the + class vectors are clustered. In the lower left part of figure \ref{fig:conv1} they are already well separated from other classes, although the hyperplane separating the o class vectors (clustered in the (0,1) corner) is still too close to the + class vectors. This is clear because 4 of these vectors have images close to the (0,1) corner. Further training should shift decision border for the o class vectors further away from the + class vectors.

The stochastic training algorithm changes network parameters along quite different trajectories in the parameter space, creating during learning very different networks, as is evident from the left and right subfigures of Fig. \ref{fig:conv1}. After some initializations convergence is very fast, with emerging separation of vectors from different classes (left subfigures of Fig. \ref{fig:conv1}. Sometimes the network gets stuck in a local minimum and inspection of the corresponding image will help to understand the problem. The lower right subfigure of Fig. \ref{fig:conv1} shows that vectors from the x class are well separated, but vectors from the two other classes have images close to the center of the triangle, extending into the lower part of the hexagon in Fig. \ref{fig:poly3}. Evidently in the feature space data vectors from these two classes are covered by the sigmoidal functions with values close to 1. Instead of waiting for the learning algorithm to correct that problem (since gradients of saturated sigmoidal functions are small this would be slow), a few simple remedies may be applied: re-initializing the network, decreasing all network parameters to make the sigmoidal functions less saturated, or perturbing the weights by adding random numbers.
Fig. \ref{fig:conv1} suggests another possibility: present as input only those vectors that correspond to images near the middle of $(a,a,0)$ line (Fig. \ref{fig:poly3}), since the network response is then closer to 0.5 than to 0 or 1, therefore gradients are relatively large and learning may proceed faster, until the scatterogram becomes more like that on the left side of Fig. \ref{fig:conv1}.

\begin{figure}
\centering
\epsfig{figure=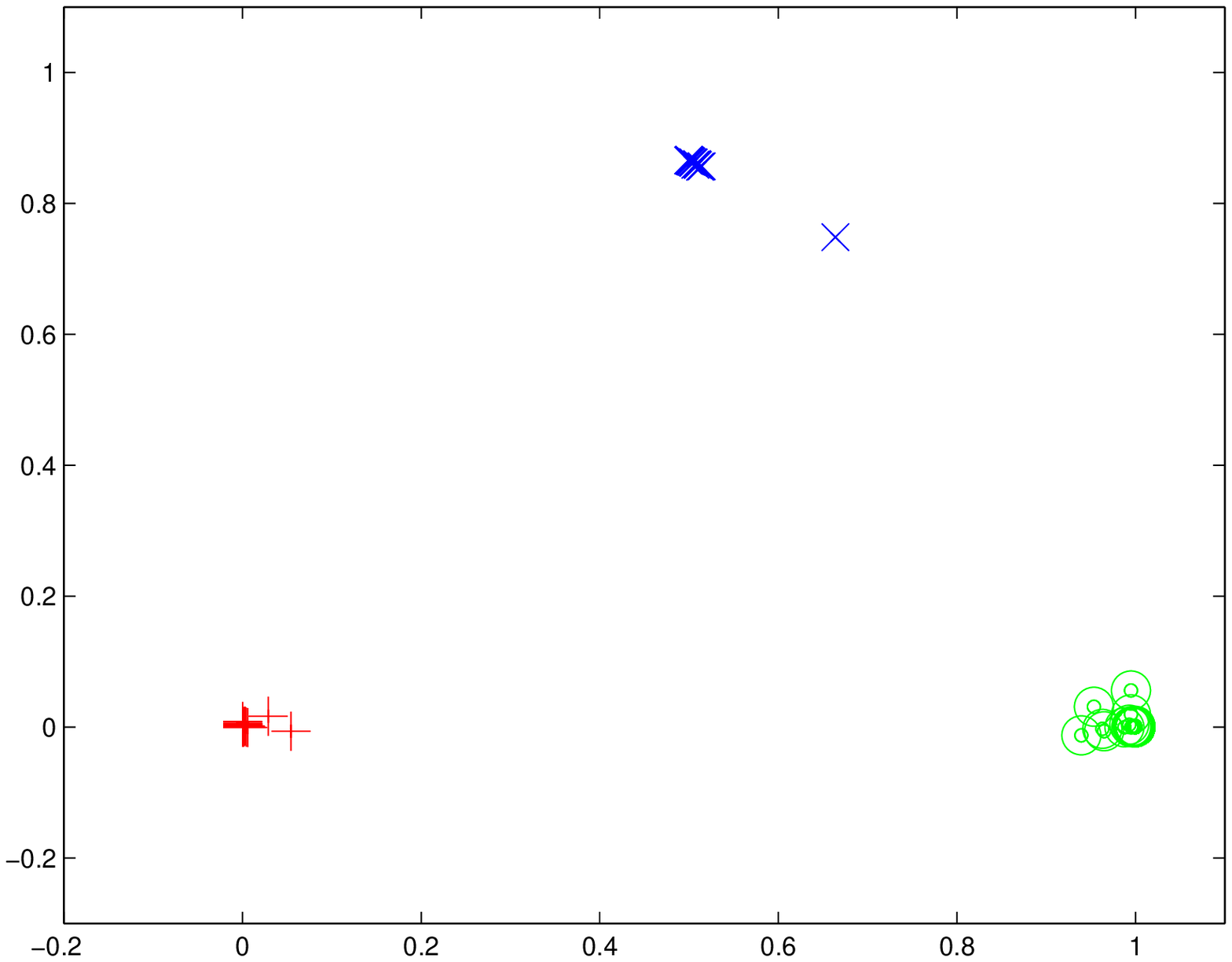,width=4.3cm}
\epsfig{figure=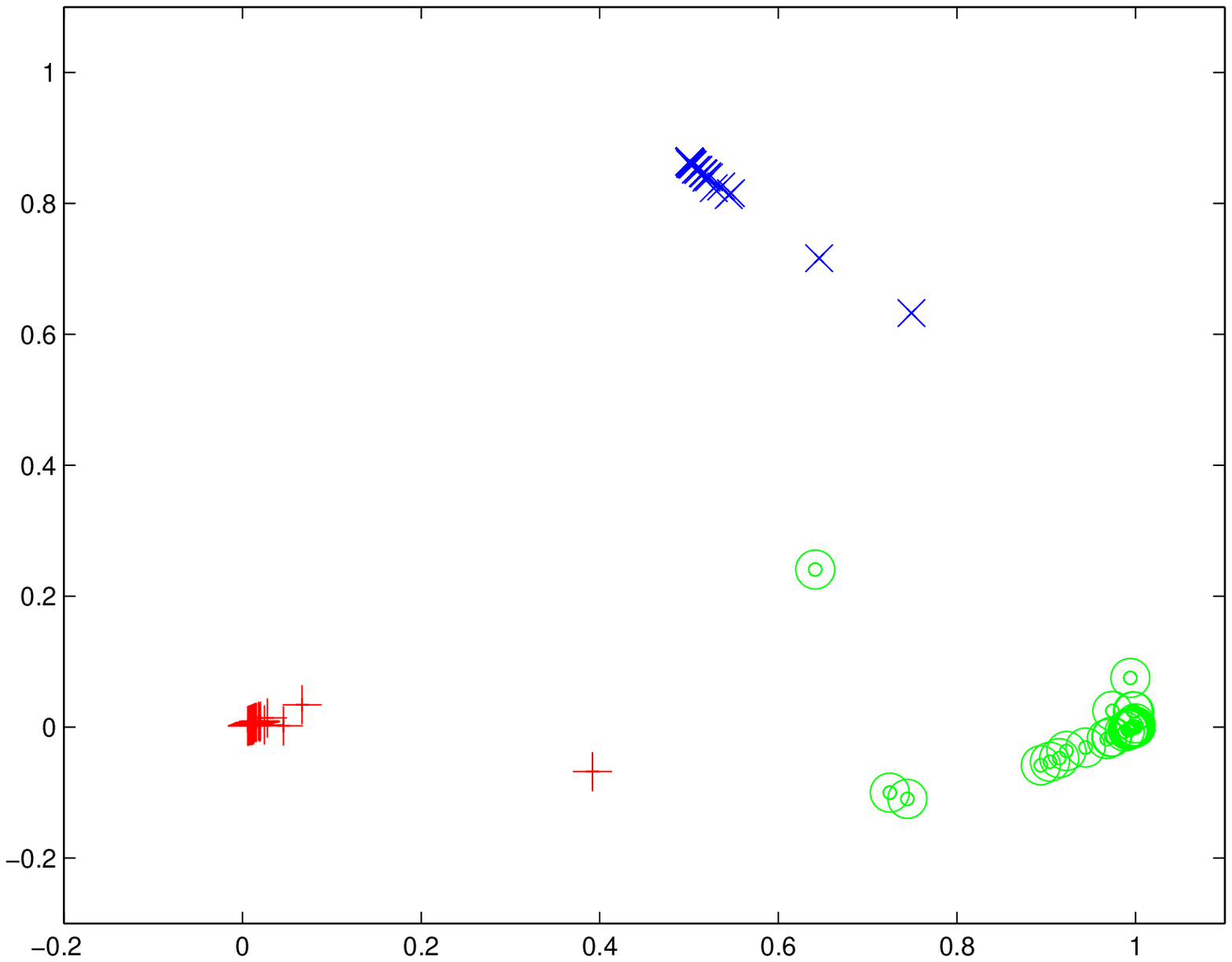,width=4.3cm}
\caption{Two converged solutions with zero errors after 30 iterations.}
\label{fig:conv2}
\end{figure}

The final solutions may look similar, although the network weights significantly differ. The size of the network weights is reflected in concentration of vector images around the corners; at the end of training (Fig. \ref{fig:conv2}) all images of training vectors cluster almost exactly in polygon's corners, indicating that the binary target values for the classes have been achieved. The number of errors is not a good indicator of the quality of solutions: both networks that were used to create Fig. \ref{fig:conv2} plots made no errors on the training data, but test results for the second network are significantly worse, since new data vectors close to the isolated + and o class vectors lead to several errors.

\subsection{Under and over-fitting effects}

\begin{figure}
\centering
\epsfig{figure=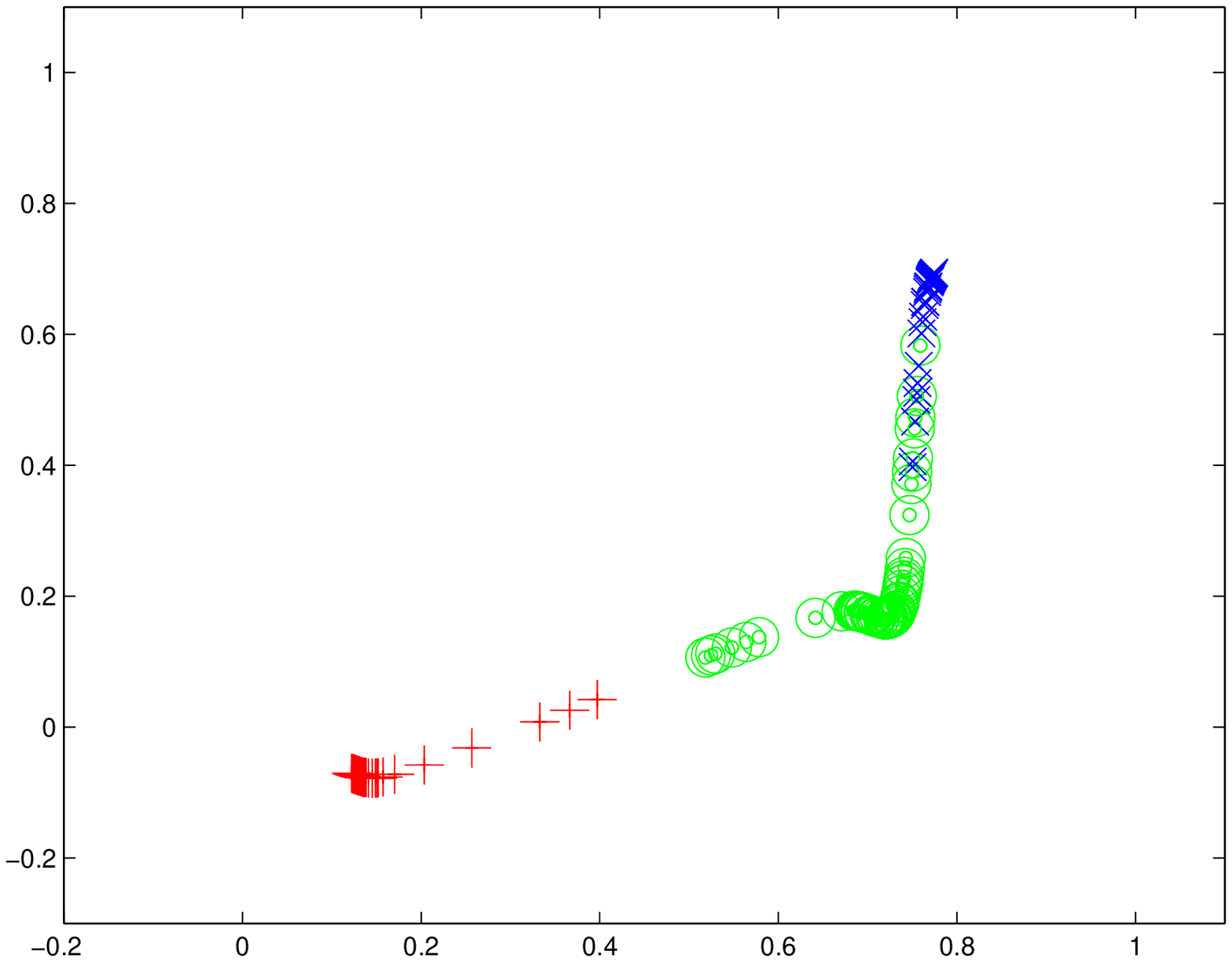,width=4.3cm}
\epsfig{figure=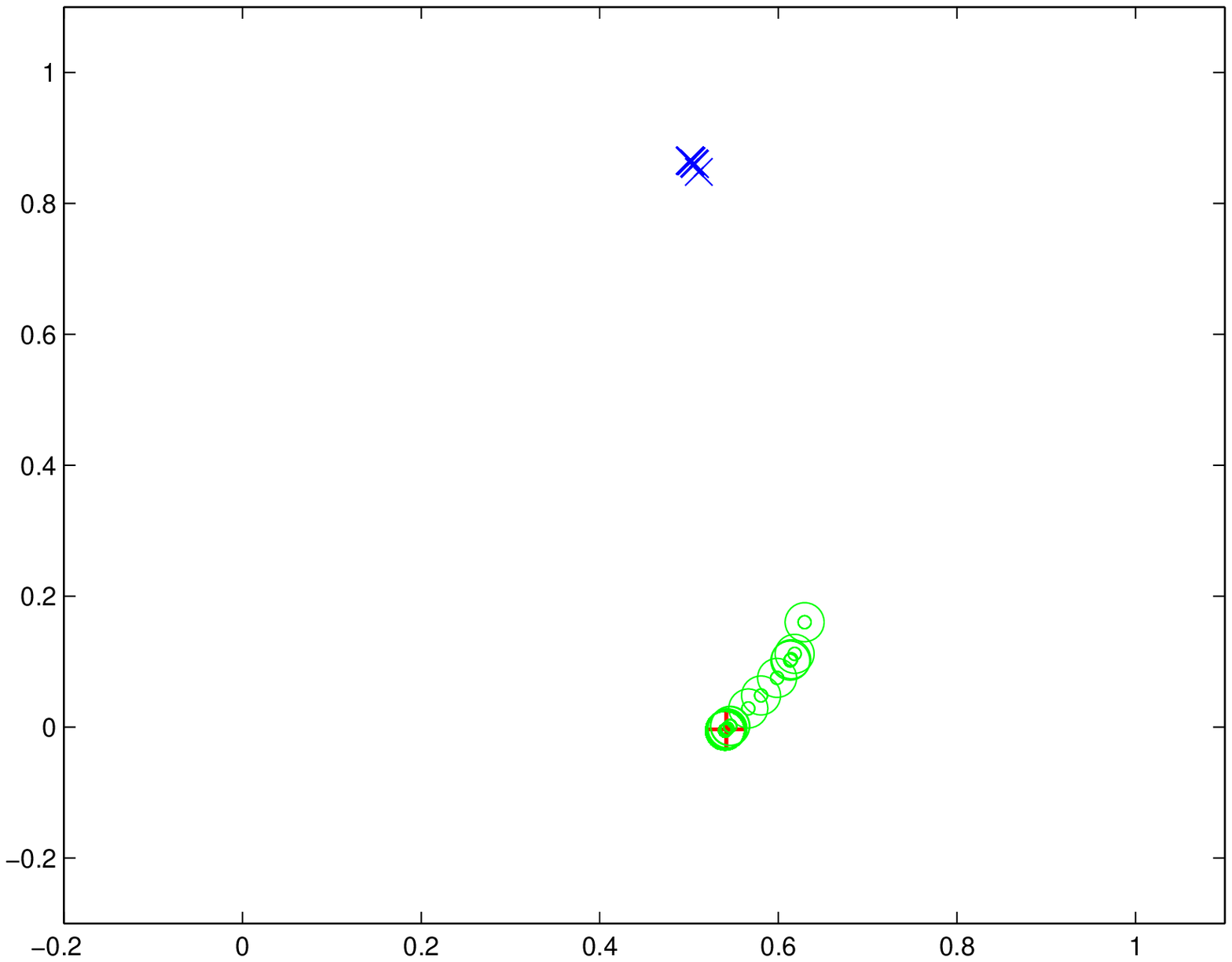,width=4.3cm}
\caption{Two converged solutions with too simple network.}
\label{fig:under}
\end{figure}

Large number of errors may result from problems with convergence -- for the Wine data some networks collapse images of all vectors into one cluster, evidently becoming trapped in a local minimum corresponding to a majority classifier. In such a case repeating the network training several times will lead to a better solution. The problem may also be due to the underfitting of the data, in which case repeating the calculation will not help. In classification problems this underfitting manifests itself with the inability of the network to create appropriate decision borders. Images of the training vectors in the scatterograms will not be clustered around the polygon vertices. In Fig. \ref{fig:under} images created by two networks with one hidden neuron are shown, one corresponding to a quite good solution with 6 errors only, and the other to a rather poor solution with 59 errors. In both cases images from one class appear in the triangle corner, while images from the two other classes appear somewhere in the middle of the triangle, showing the inability of the network to find a proper solution.

\begin{figure}
\centering
\epsfig{figure=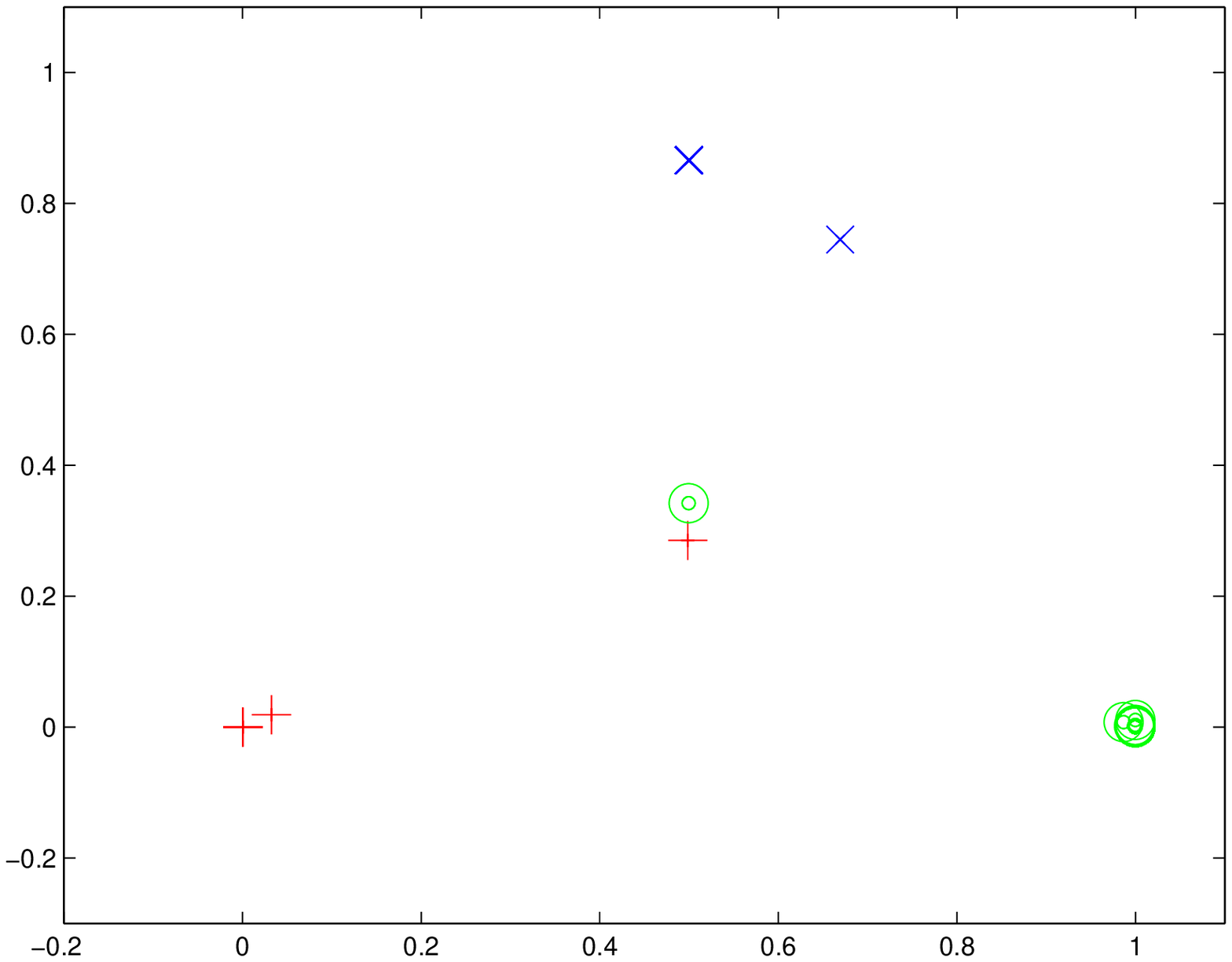,width=4.3cm}
\epsfig{figure=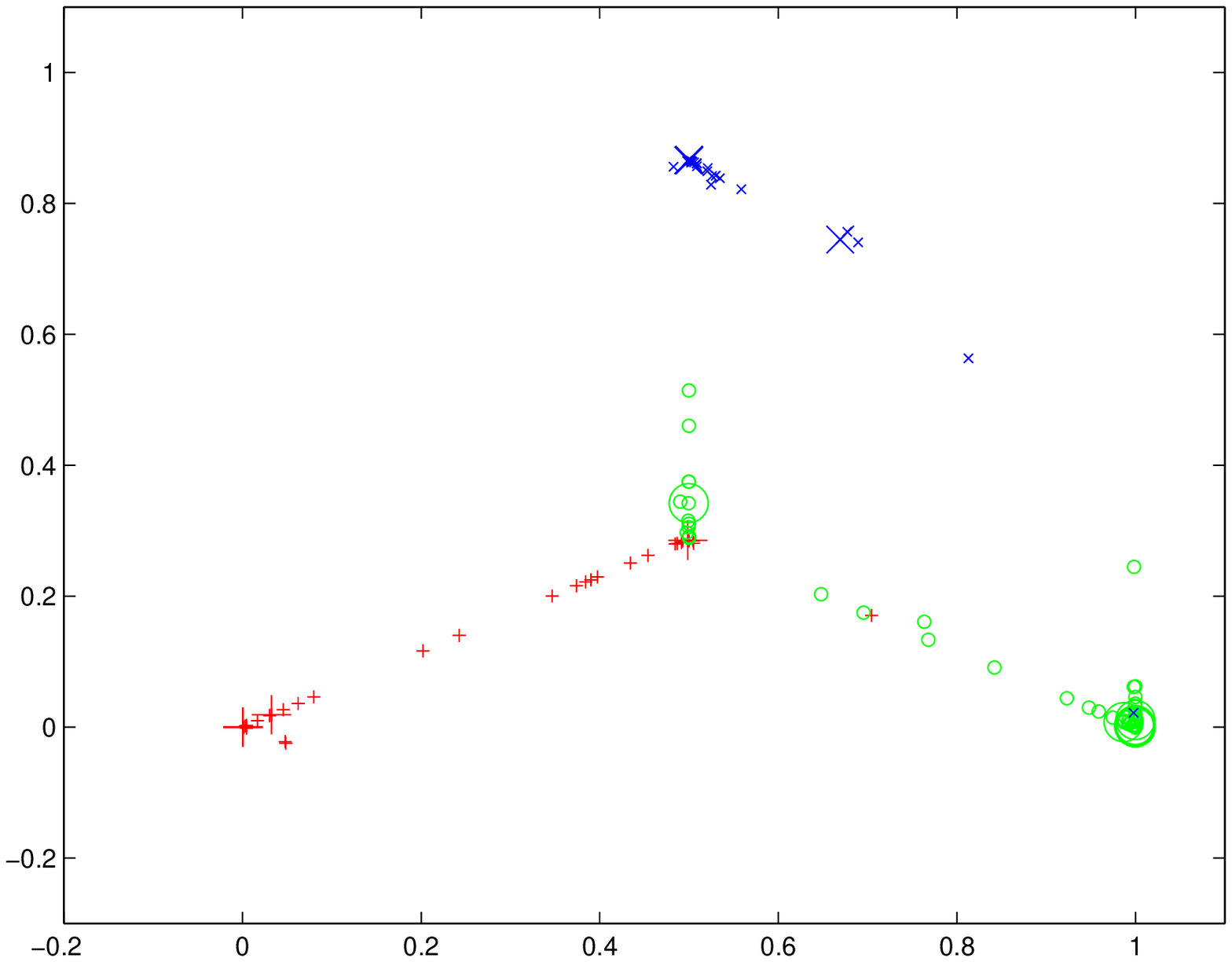,width=4.3cm}
\caption{Left plot -- mapping by an MLP network that is too complex; right side -- the same mapping applied to more vectors, created by adding small variance noise to the training vectors.}
\label{fig:over}
\end{figure}

On the other hand networks may be too complex, overfitting the data. Training of the MLP network with 30 hidden neurons has been done on 2/3 of the randomly selected data, and results are displayed for all data. Although no errors have been made on the training partition, images of several test vectors appear near the center of the triangle, corresponding to vectors that the network does not recognize (all network outputs are quite small), indicating that the network does not generalize well. This is confirmed by adding noise to original data -- in Fig. \ref{fig:over} small x, o and + are images of original data vectors, slightly perturbed with Gaussian distributed random vectors of unit variance multiplied by 0.02. The lines between the center and the triangle vertices show that some perturbed vectors are in regions of the feature space where all sigmoidal functions of the MLP network have small values.

\subsection{Regularization effects}

\begin{figure}
\centering
\epsfig{figure=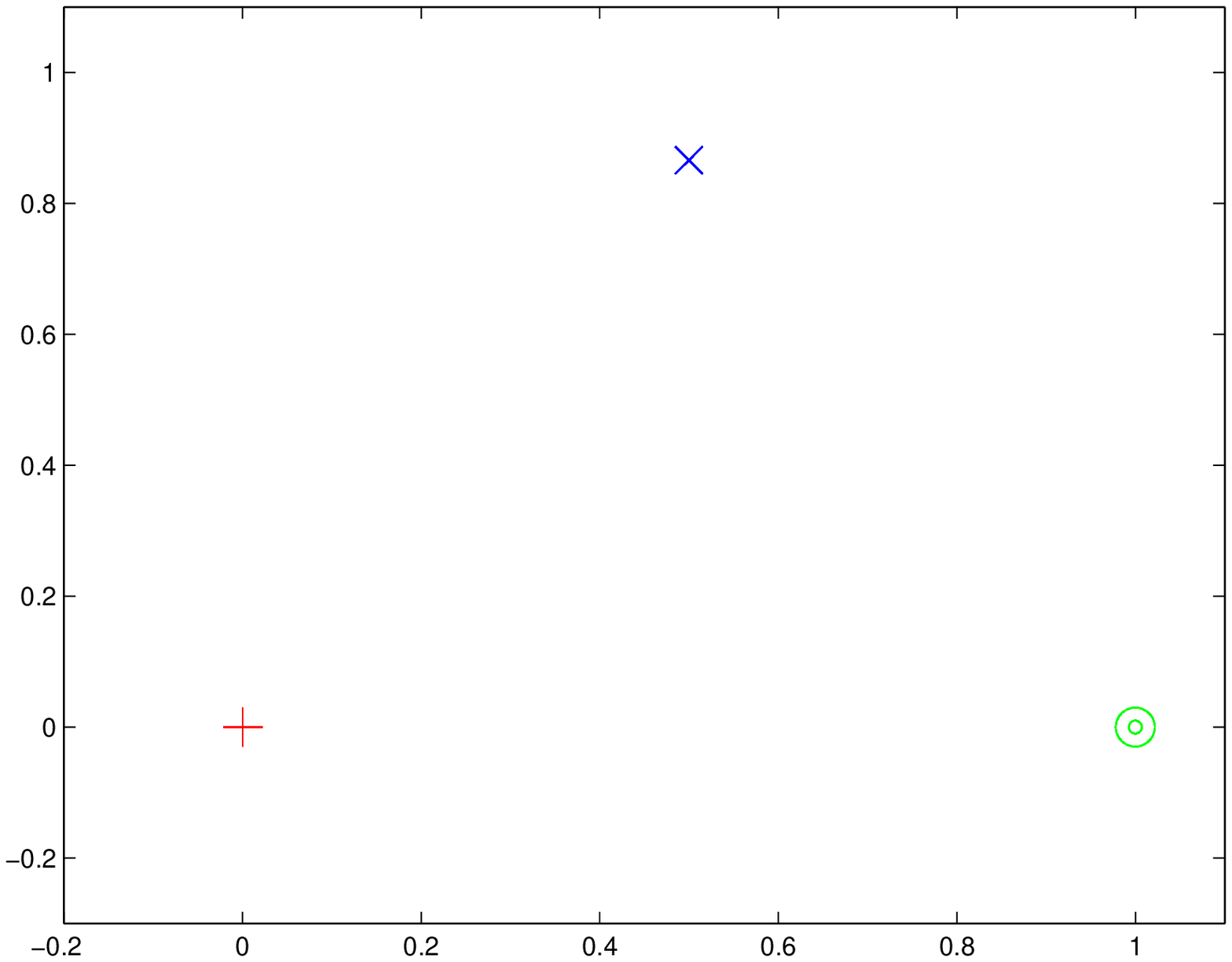,width=4.3cm}
\epsfig{figure=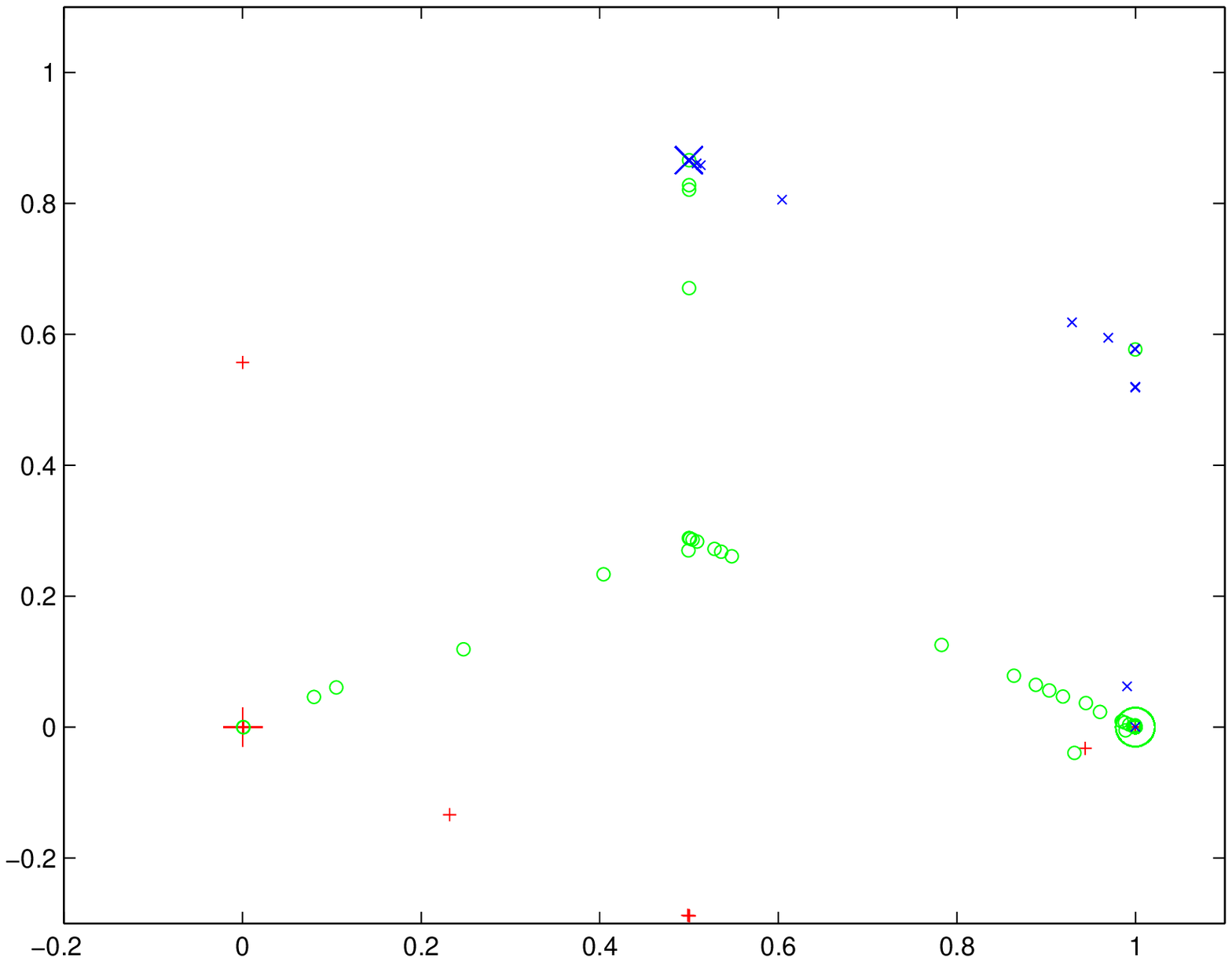,width=4.3cm}
\epsfig{figure=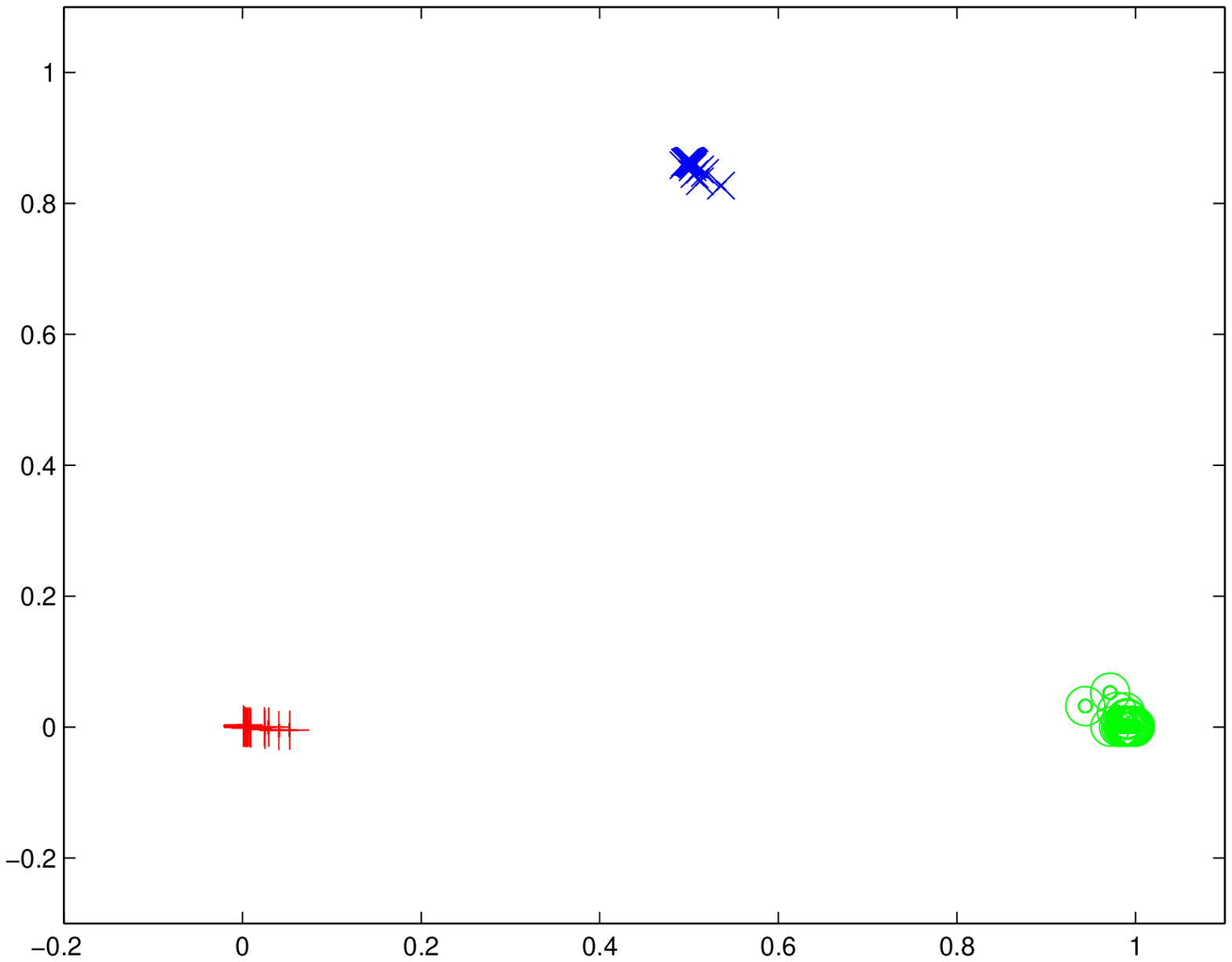,width=4.3cm}
\epsfig{figure=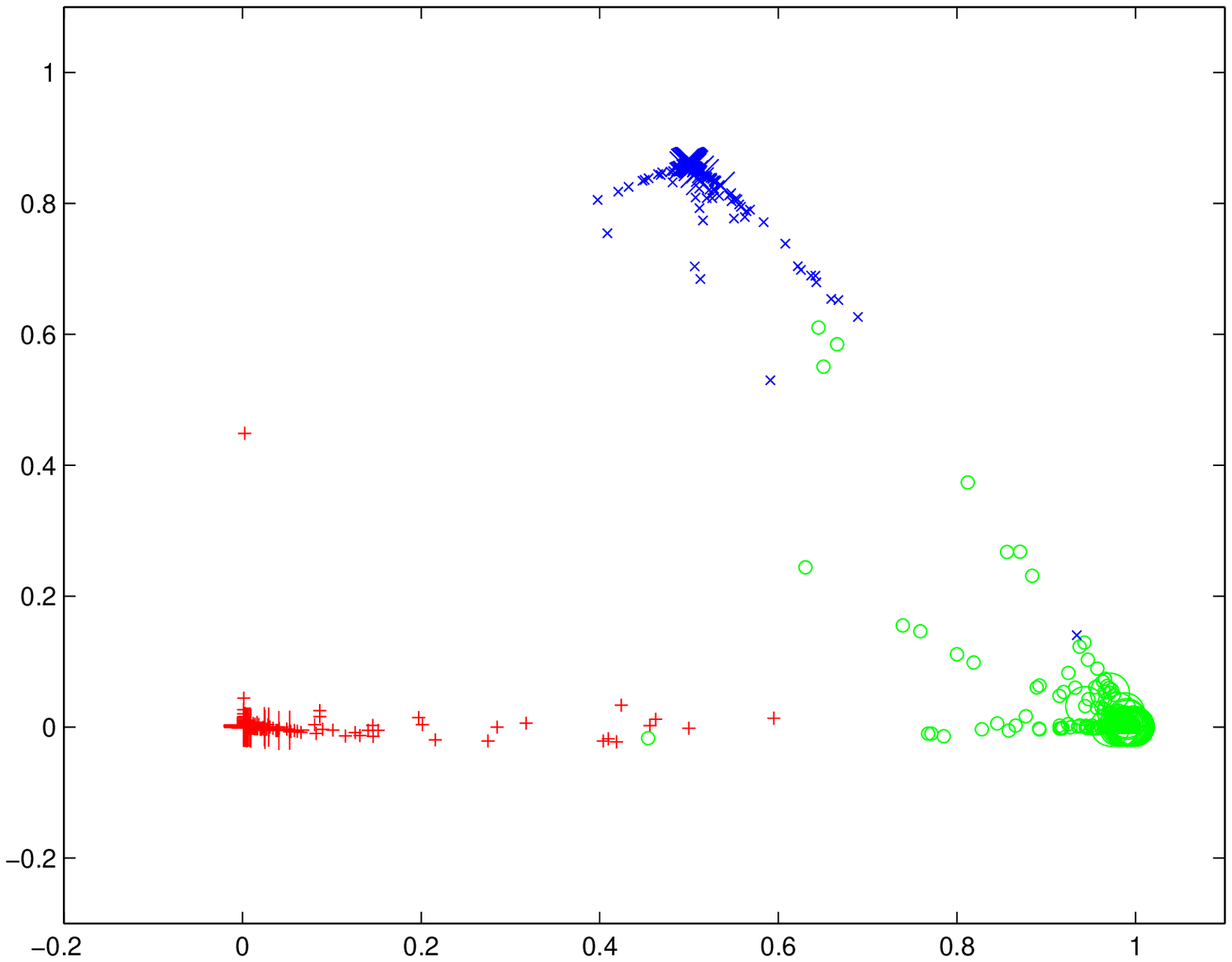,width=4.3cm}
\epsfig{figure=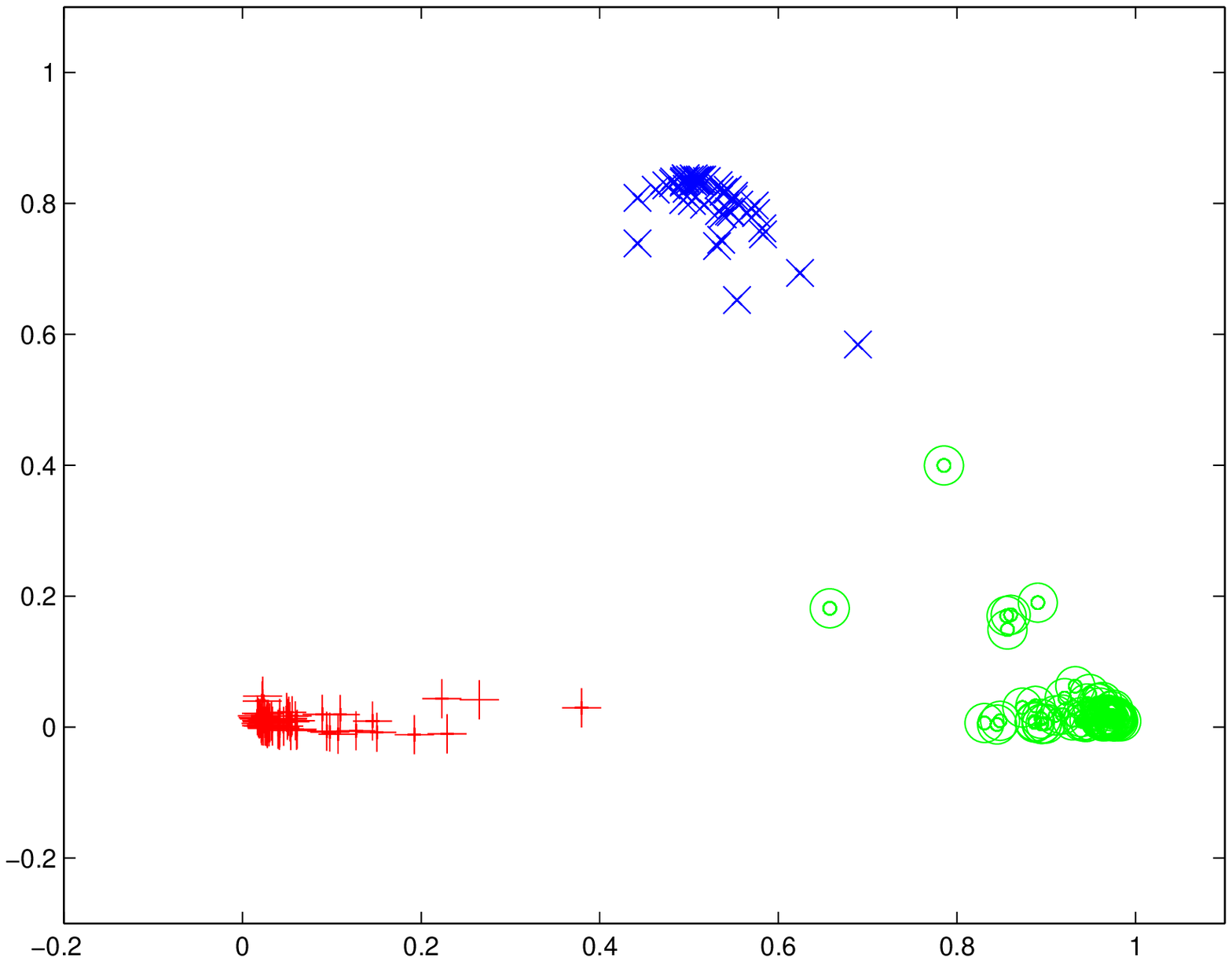,width=4.3cm}
\epsfig{figure=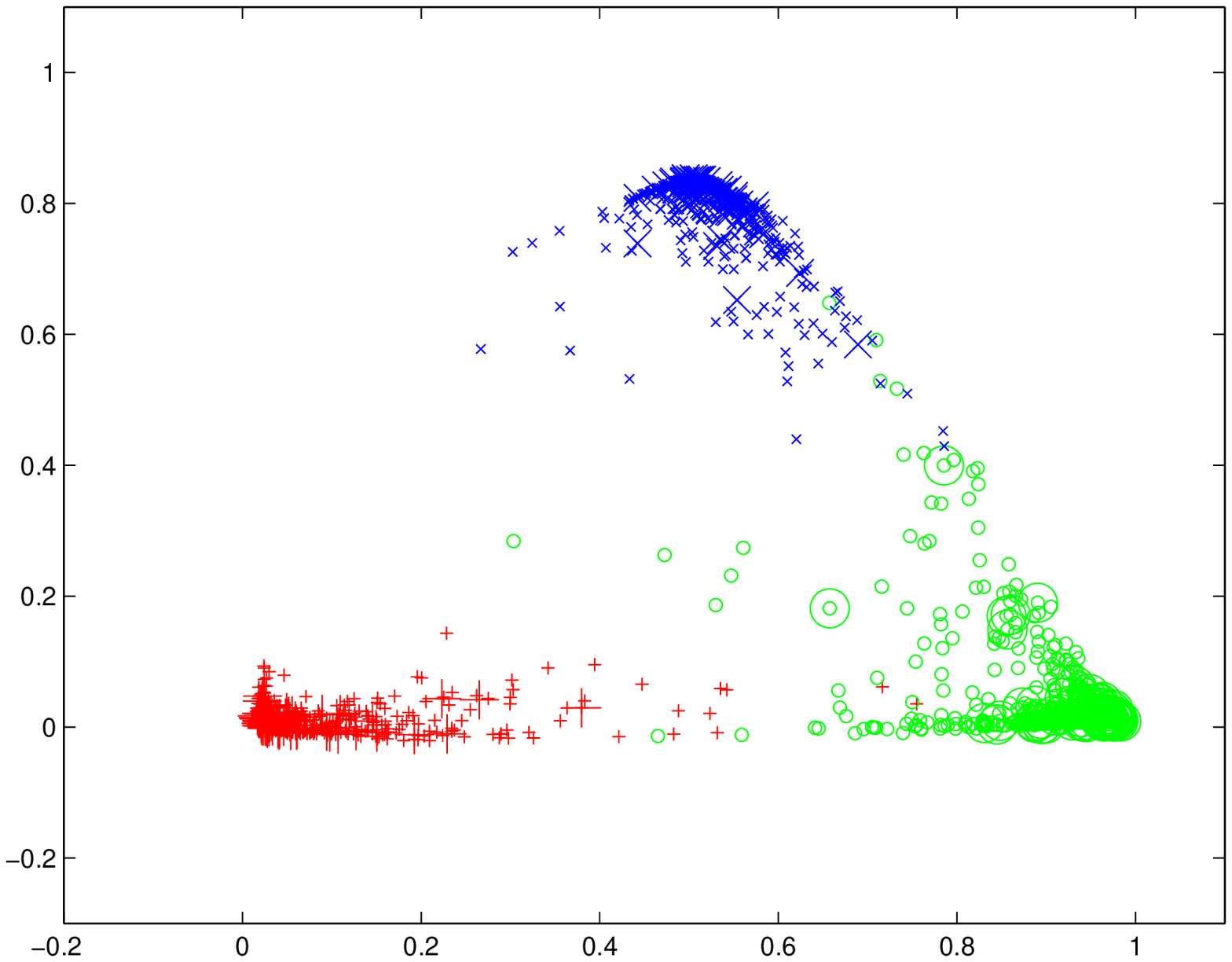,width=4.3cm}
\epsfig{figure=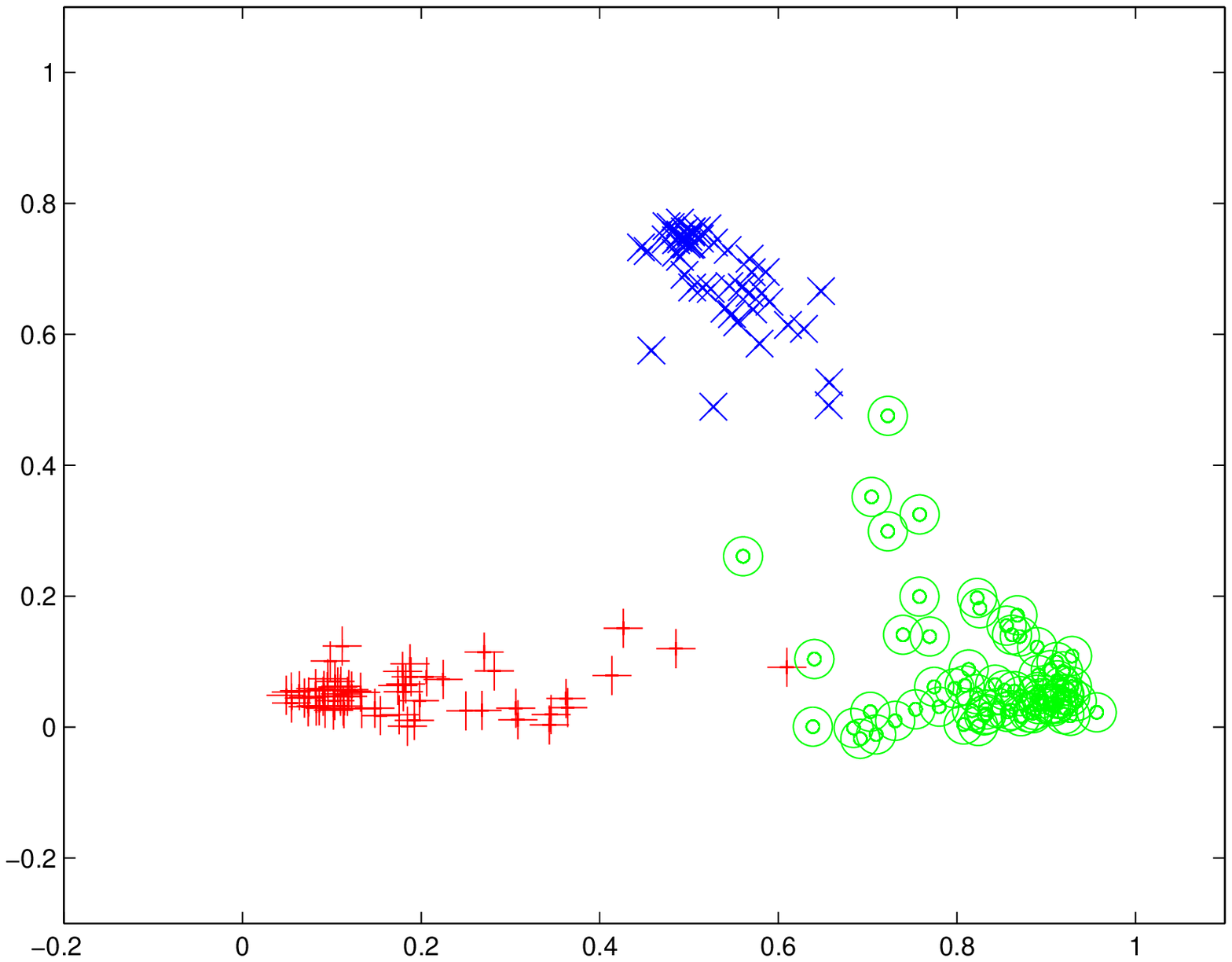,width=4.3cm}
\epsfig{figure=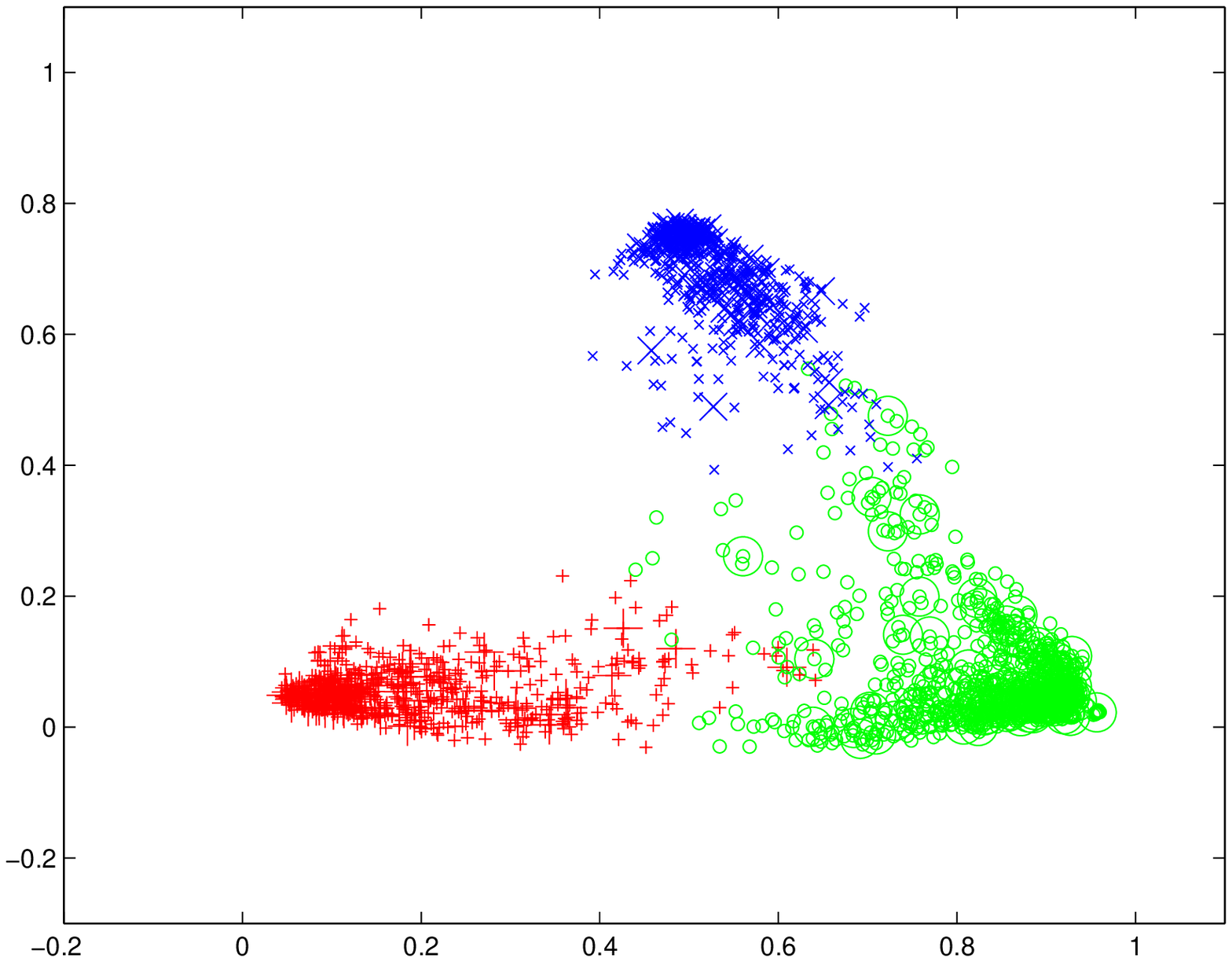,width=4.3cm}
\caption{Effects of regularization: top row - no regularization, second row $\alpha=0.05$, third row $\alpha=1.0$, and fourth row $\alpha=5.0$. Figures on the right side are with 5\% of Gaussian noise.}
\label{fig:reg}
\end{figure}

After convergence images of the training vectors may collapse into a single point, showing that the network is over-confident, and the images of vectors that are classified wrongly will be mapped into wrong vertices of the polygon. MLP networks behave in this way when weights become very large, creating almost step-like functions that correspond to sharp decision borders. Such decision borders may be brittle, and will lead to poor generalization of the network. Perturbing training vectors by adding some noise will show this effect clearly in scatterograms -- lines connecting vertices with the polygon's center will appear, as in the right plot in Fig. \ref{fig:over}, and the top right plot in Fig. \ref{fig:reg}. In fact adding noise to the input data is equivalent to a regularization procedure \cite{Bishop}, making the solutions more robust and increasing classification margins.

Wide margin solutions are manifested by images of the training vectors concentrated near polygon vertices, but not collapsed into a single point. The network is not overconfident, i.e. the errors are closer to the center of the polygon or close to the midpoints of lines connecting polygon's vertices.
This is shown in Fig. \ref{fig:reg} for network with 3 hidden units that was able to perfectly separate the training data. Without regularization images of the training vectors generated by the network collapse into three vertices of the triangle, while images of some perturbed vectors (5\% Gaussian noise) lie on the line joining vertices with centers, indicating that these vectors are in the region where no sigmoidal function has a large value (Fig. \ref{fig:reg}, top row). Gaussian regularization prior added to the MLP error function scaled by a small $\alpha=0.05$ hyperparameter partially removes this effect, making the corners more blurred and removing images of the perturbed vectors from the center, although the images of the training vectors are still very close to the triangle vertices. Increasing the regularization hyperparameter to $\alpha=1.0$ and $\alpha=5.0$ makes the network much less confident and shows more realistic predictions, because some samples of wines from the + class happen to be rather similar to samples from o class, and those from the o class are similar to samples from the x class. With very large regularization hyperparameter the network will start to make some errors, but even for $\alpha=5.0$ images of almost all perturbed vectors are concentrated around correct corners of the triangle. Thus visualization may be useful to select the best network with proper regularization.

\subsection{Differences between networks of the same accuracy}

\begin{figure}
\centering
\epsfig{figure=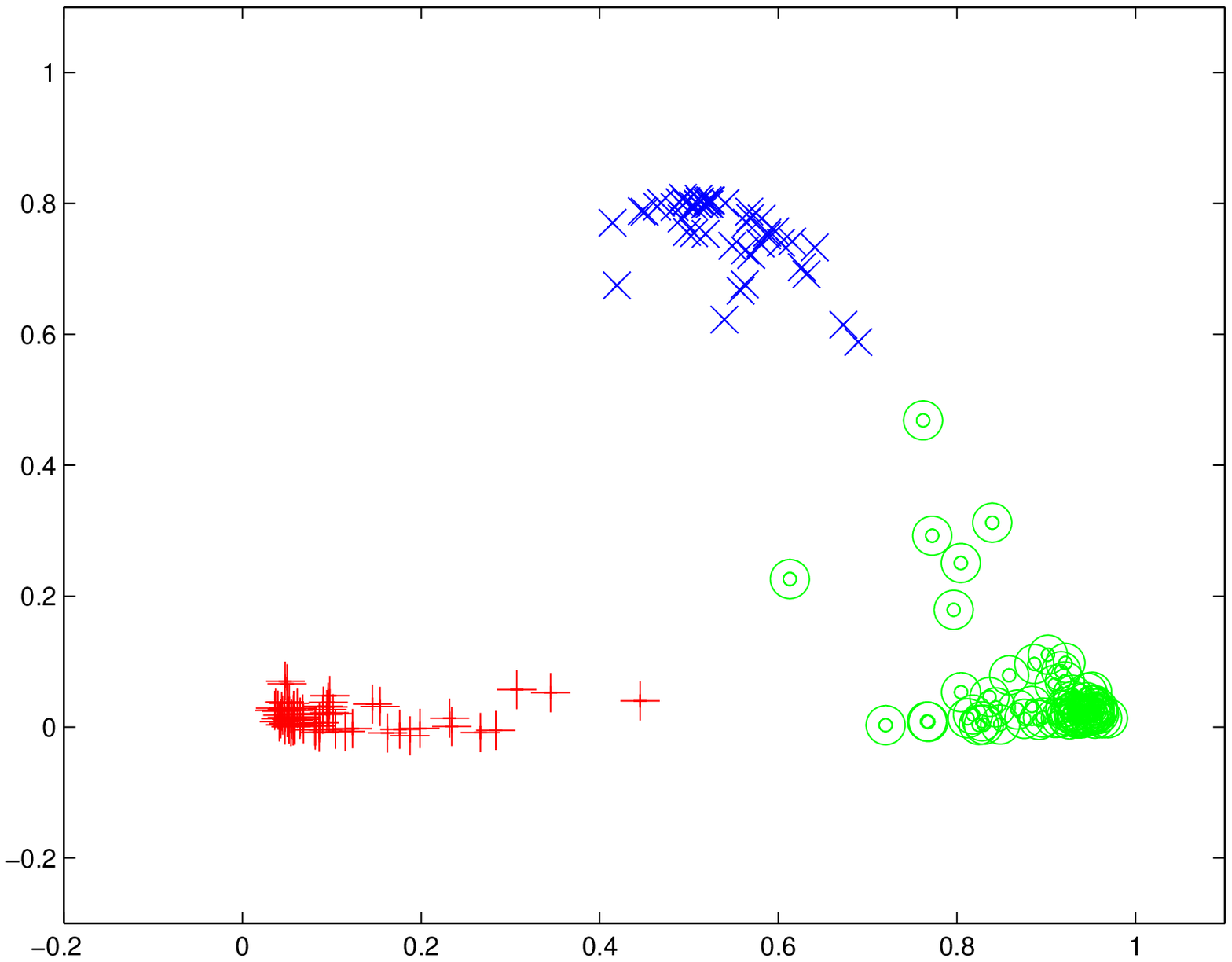,width=4.3cm}
\epsfig{figure=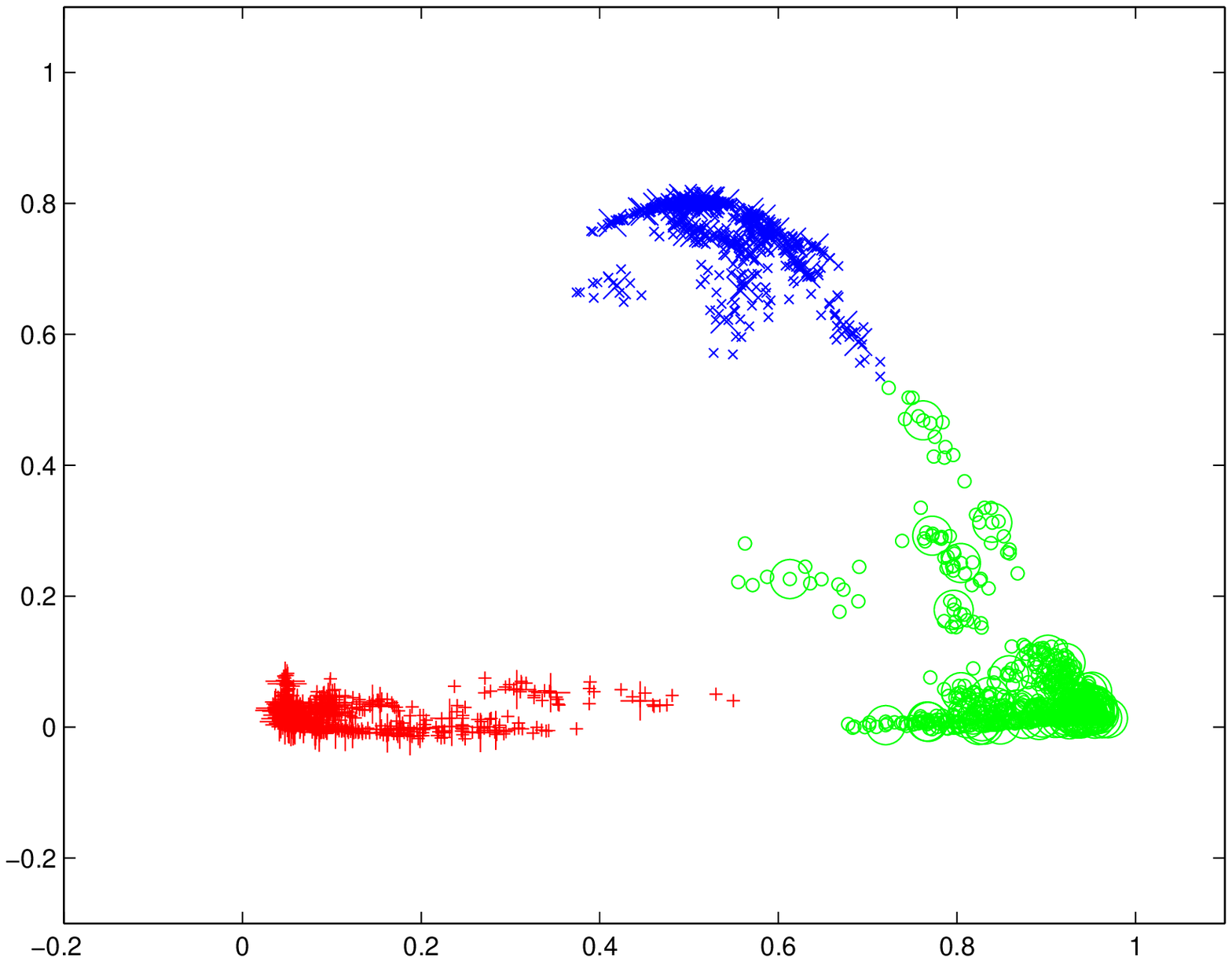,width=4.3cm}
\epsfig{figure=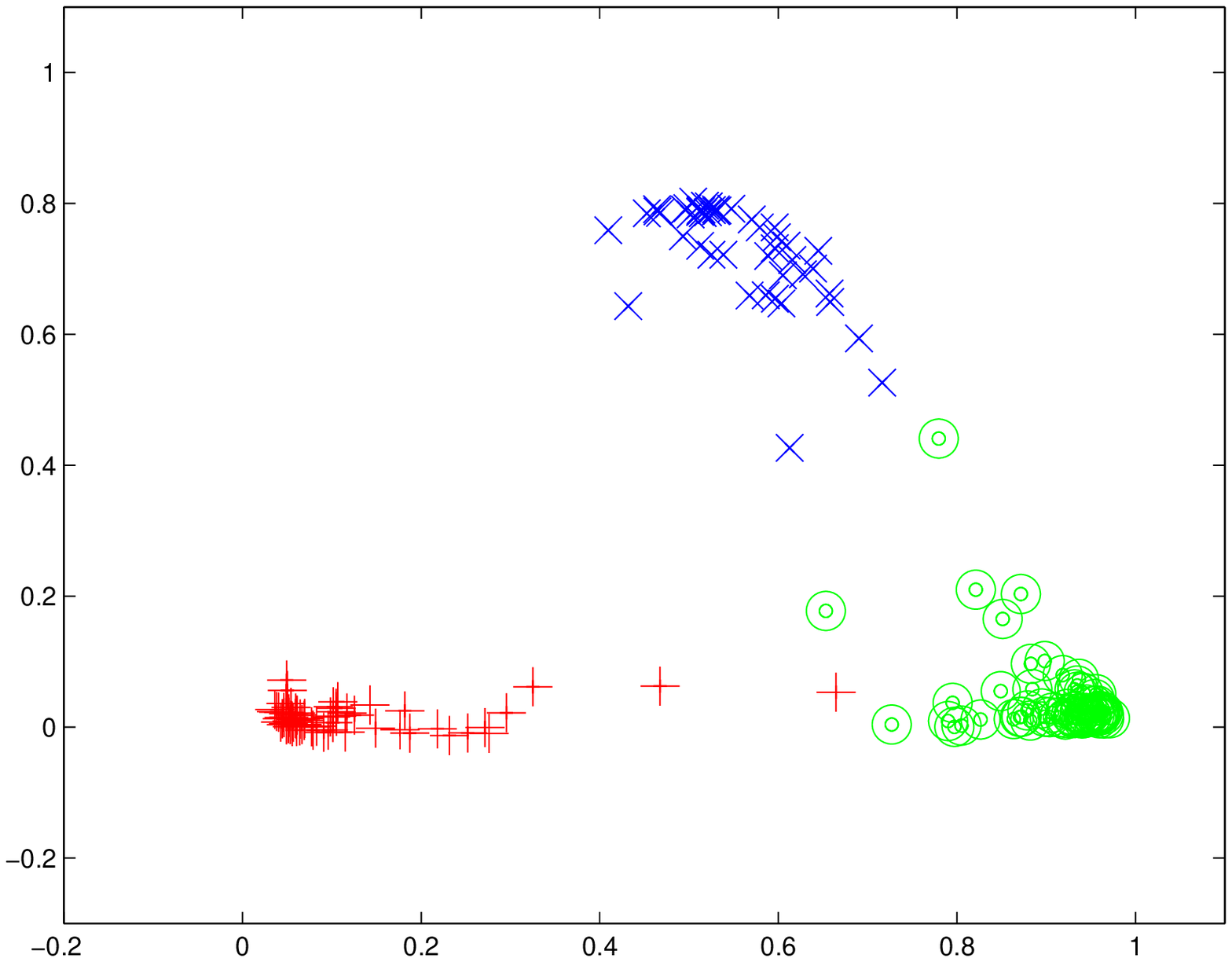,width=4.3cm}
\epsfig{figure=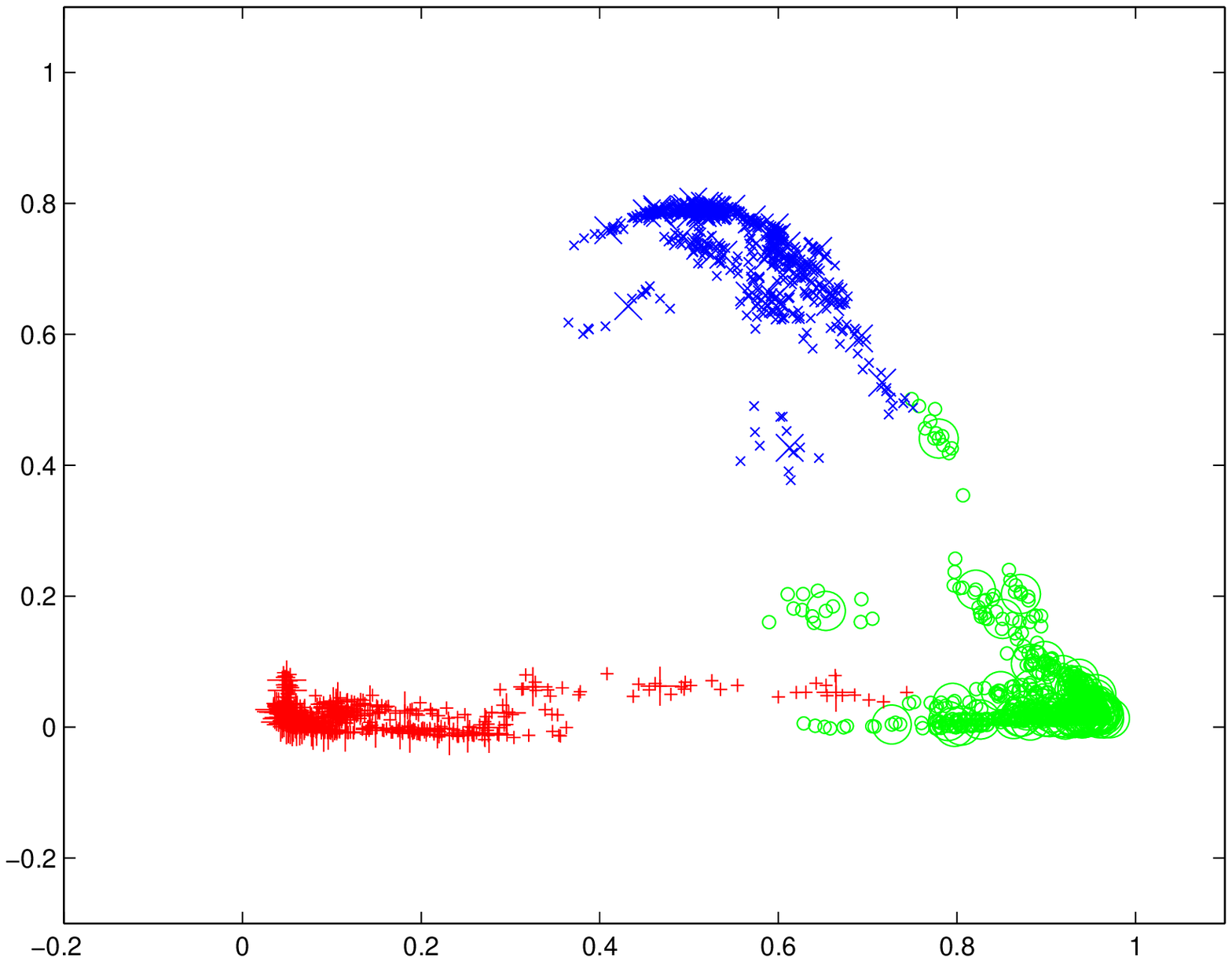,width=4.3cm}
\caption{Two networks, each making only one error on the training data; the first (top row) has higher chance to mix classes x and o more often, the second (bottom row) to mix classes + and o more often.}
\label{fig:dif}
\end{figure}

Two networks with similar MSE, making the same number of errors and having identical confusion matrices, may still significantly differ in some areas of the feature space. In the Wine example, vectors from + and o classes may be quite close to the decision surface, or vectors from x and o class may be close to the decision surface. Although in both cases same errors have been made so far, one network may be preferred over the other if the costs of mixing different classes are not equivalent. This is demonstrated in Fig. \ref{fig:dif}  by adding low variance noise (2\%) to perturb original data.

\begin{figure}
\centering
\epsfig{figure=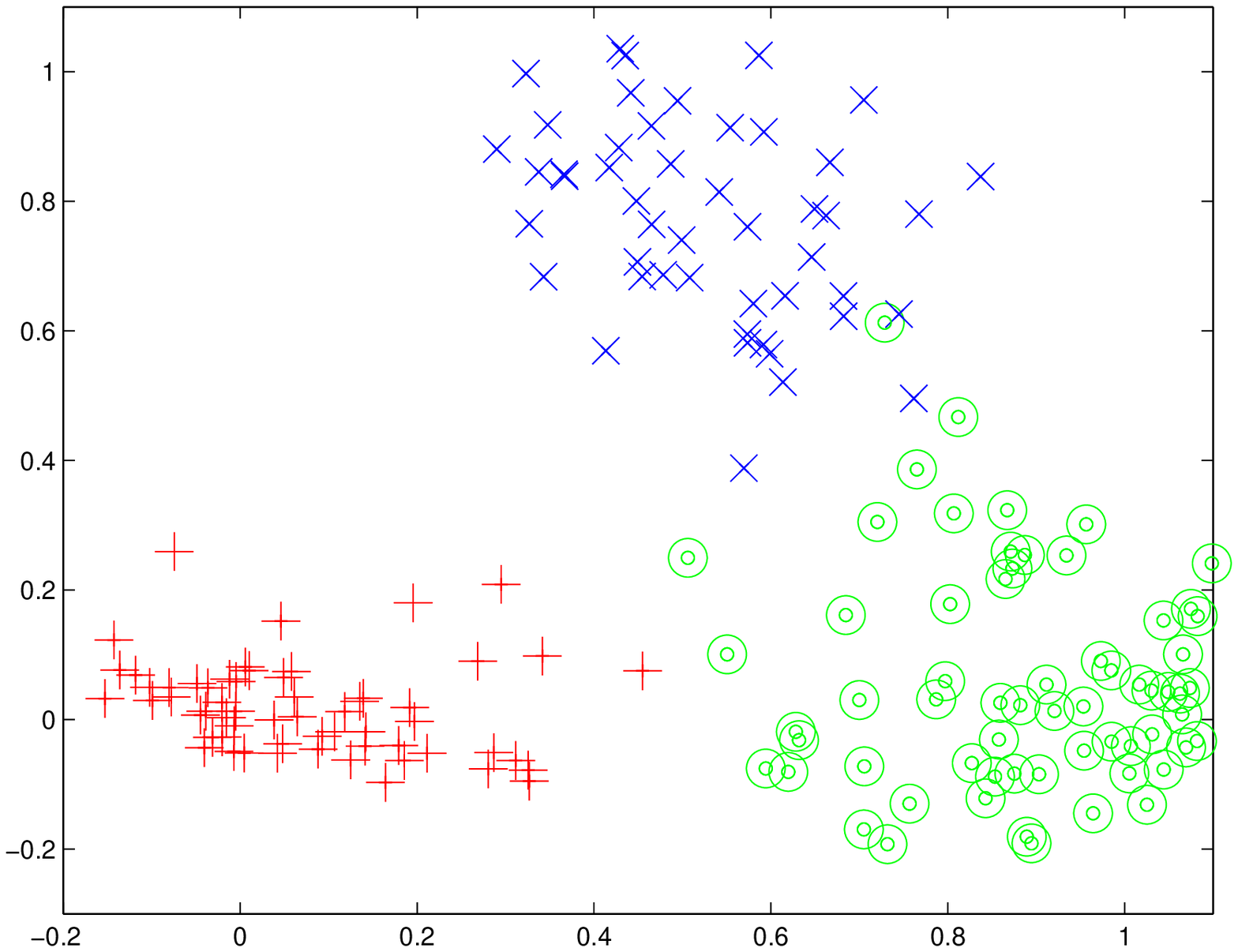,width=4.3cm}
\epsfig{figure=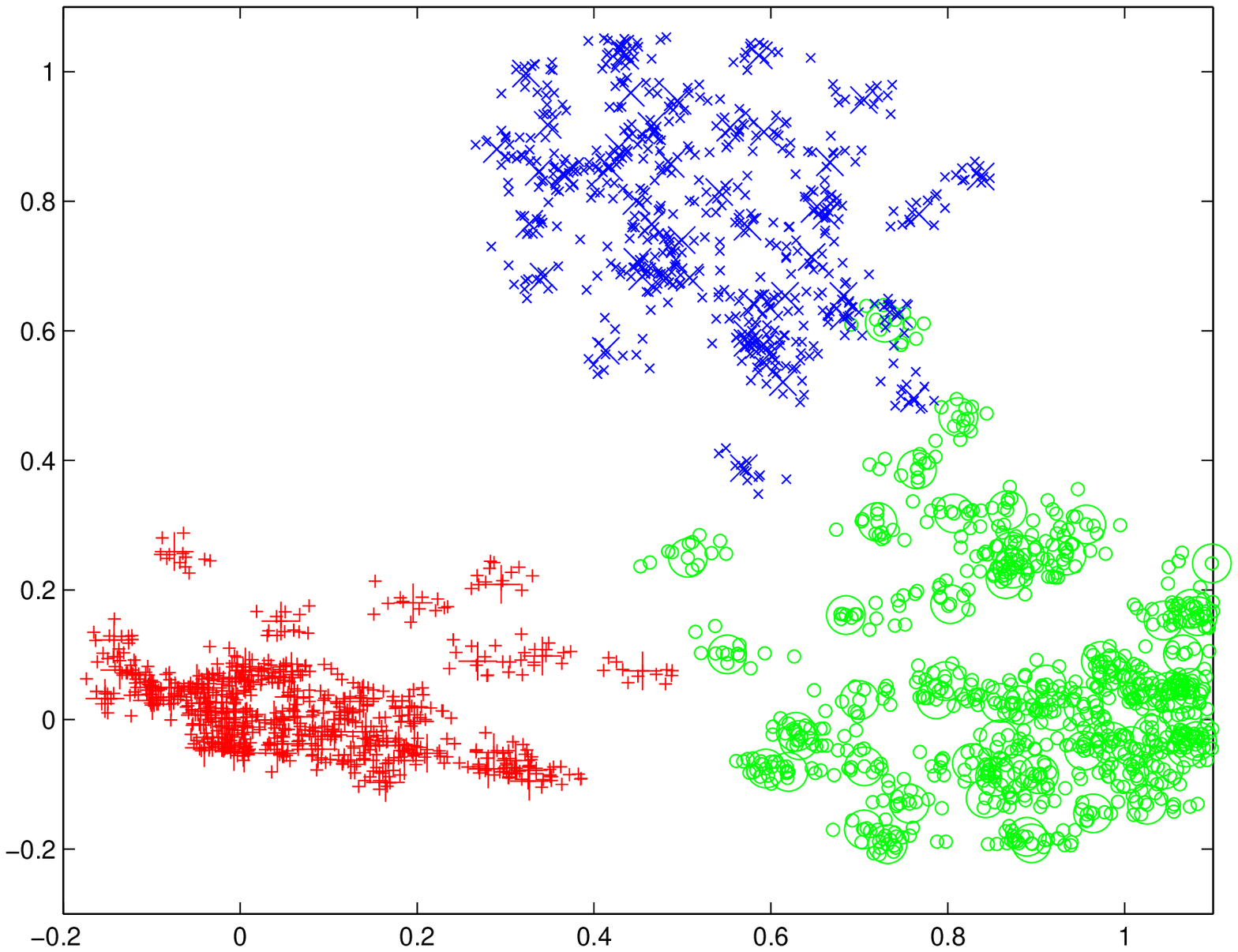,width=4.3cm}
\epsfig{figure=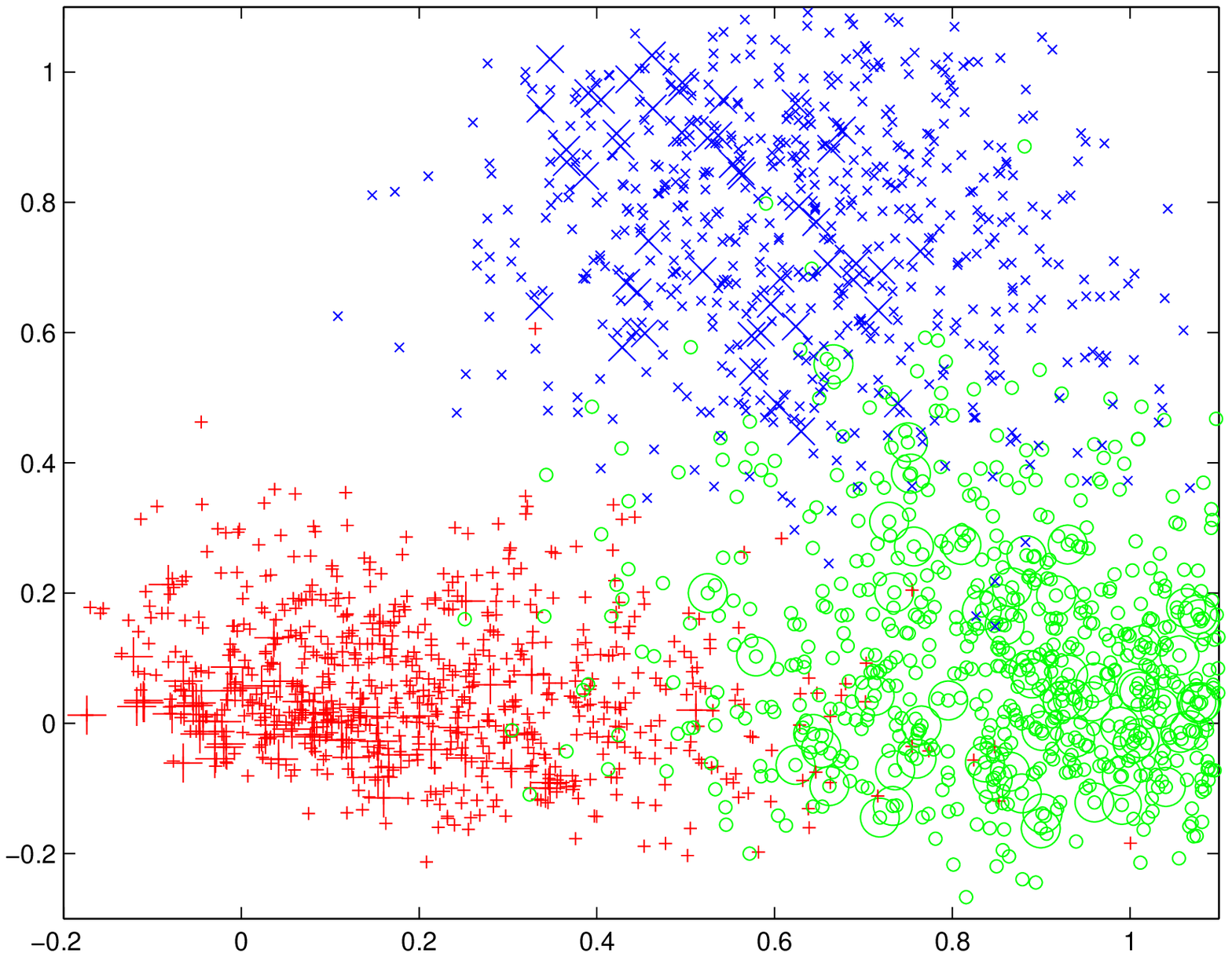,width=4.3cm}
\epsfig{figure=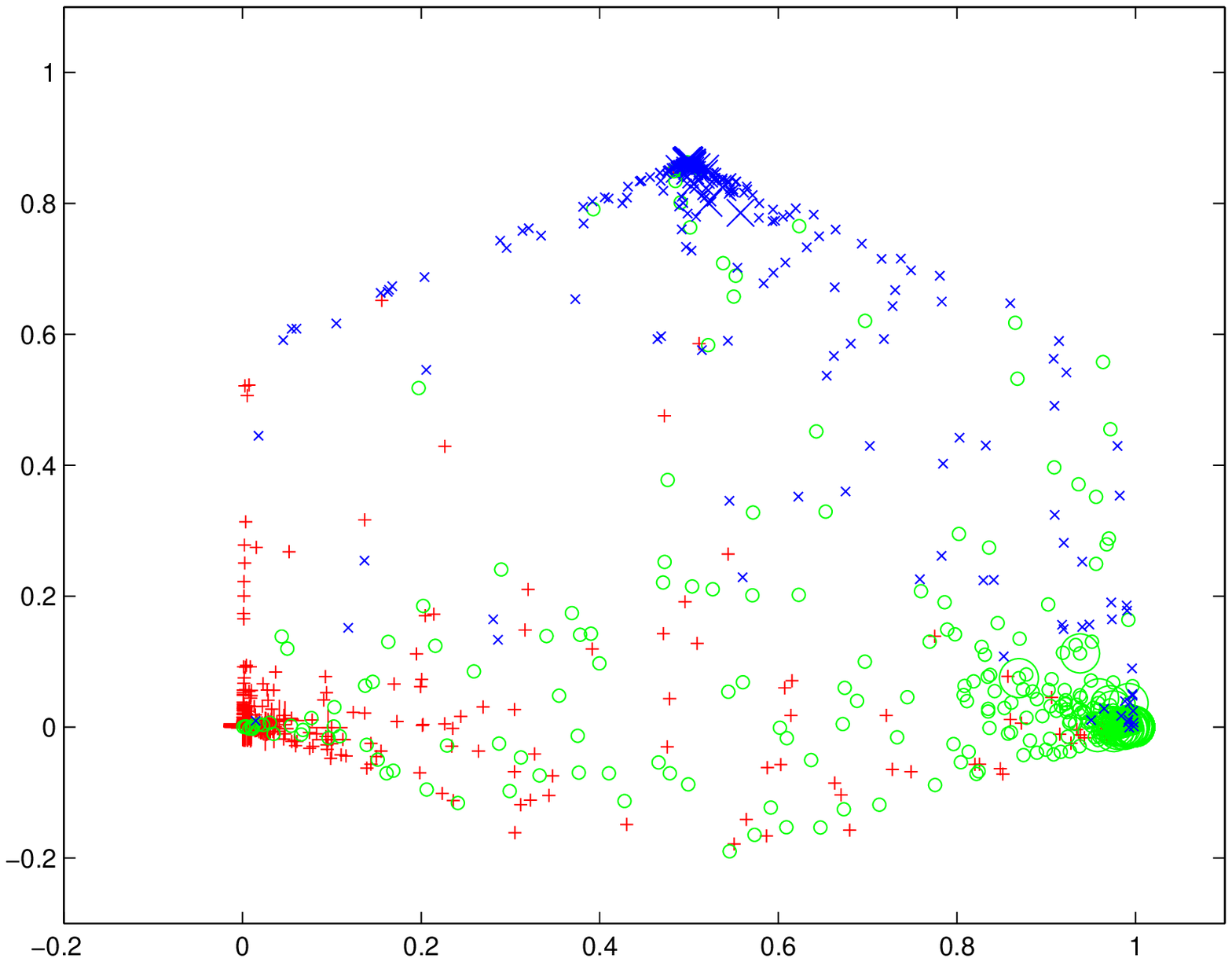,width=4.3cm}

\caption{Top row: RBF network solution with 6 Gaussian functions; right figure -- same RBF network on slightly perturbed (2\%) input vectors.
Bottom row: comparison of RBF with MLP solutions for inputs perturbed by a strong noise (15\%) .}
\label{fig:rbf}
\end{figure}

Different gradient optimization procedures will also converge to different networks. These differences are visible even better if RBF network is used instead of an MLP. With 6 Gaussian functions RBF network also finds a solution with a single error. The images of the training vectors after mapping through the RBF network are much less localized, while the perturbed vectors are much closer to the unperturbed vectors (Fig. \ref{fig:rbf}, top right) than for MLPs. Nonlinearities introduced by the RBF network are significantly smaller than those of the MLP network (especially with no regularization), therefore the RBF solution is more robust.
Perturbing original vectors with noise with large variance will not elicit any unexpected behavior from the RBF network (bottom row, Fig. \ref{fig:rbf}). MLP network with small regularization ($\alpha=0.1$) and the same number of hidden units makes less errors, but places many perturbed vectors close to vertices corresponding to wrong classes (i.e. makes erros with high confidence). Images of vectors mapped by MLP show only how close these vectors are to the decision borders, while images obtained with RBF mapping show also similarities between vectors in feature spaces.

For easy problems, with well separated clusters, MLP with regularization provides quite robust solutions.
MLP with 5 hidden neurons and strong regularization ($\alpha=1$) creates images of vectors from 5 classes, clustered in vertices of a pentagon. The network mapping is quite robust, even after adding noise with 100\% variance the network behavior is quite predictable, indicating that no strange kinks are hiding in its black box. The ``arms'' extending from one of the vertices to two other vertices
simply indicate that the feature space vectors corresponding to these images belong to clusters that are relatively close together.

%

%

\begin{figure}
\centering
\epsfig{figure=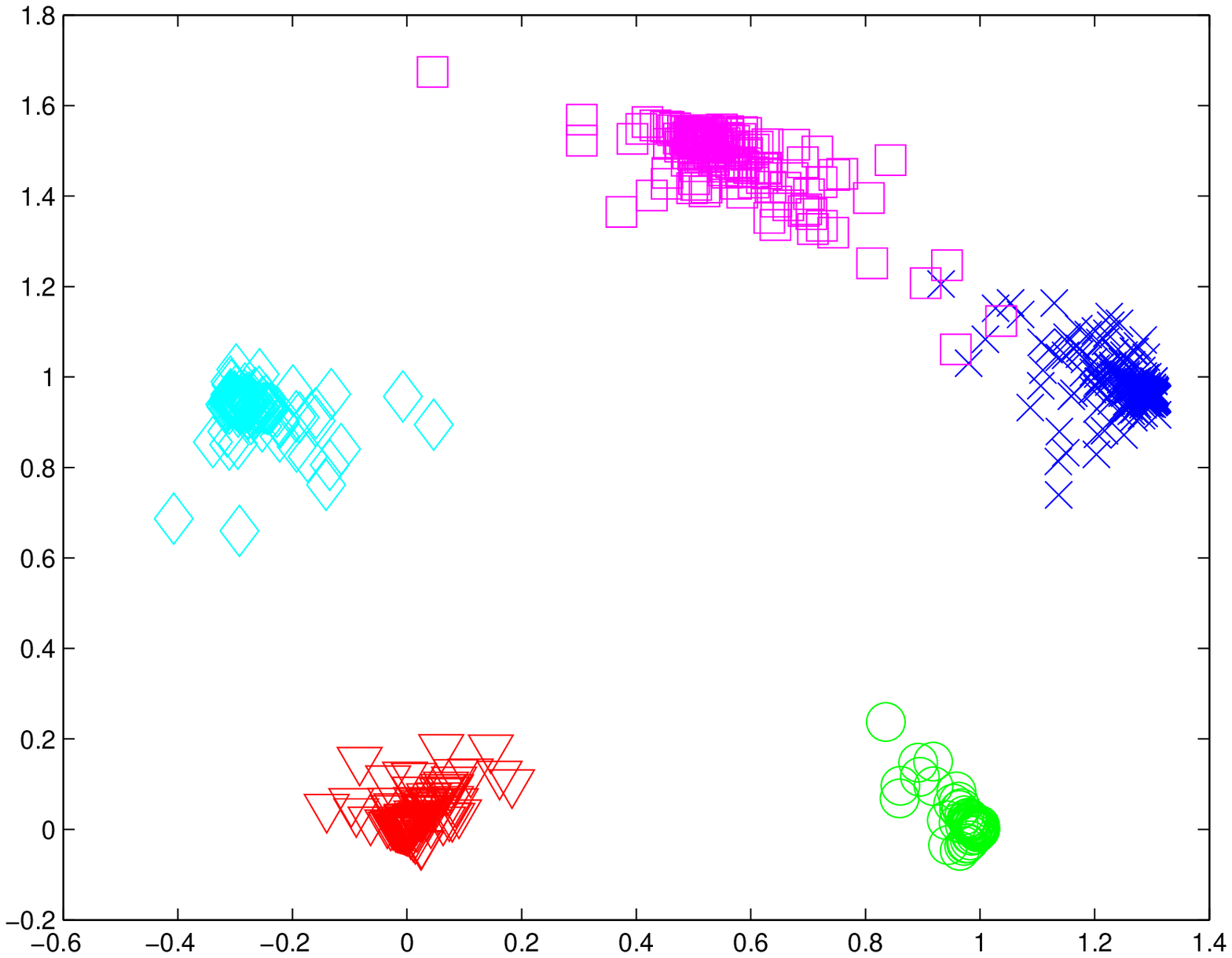,width=4.3cm}
\epsfig{figure=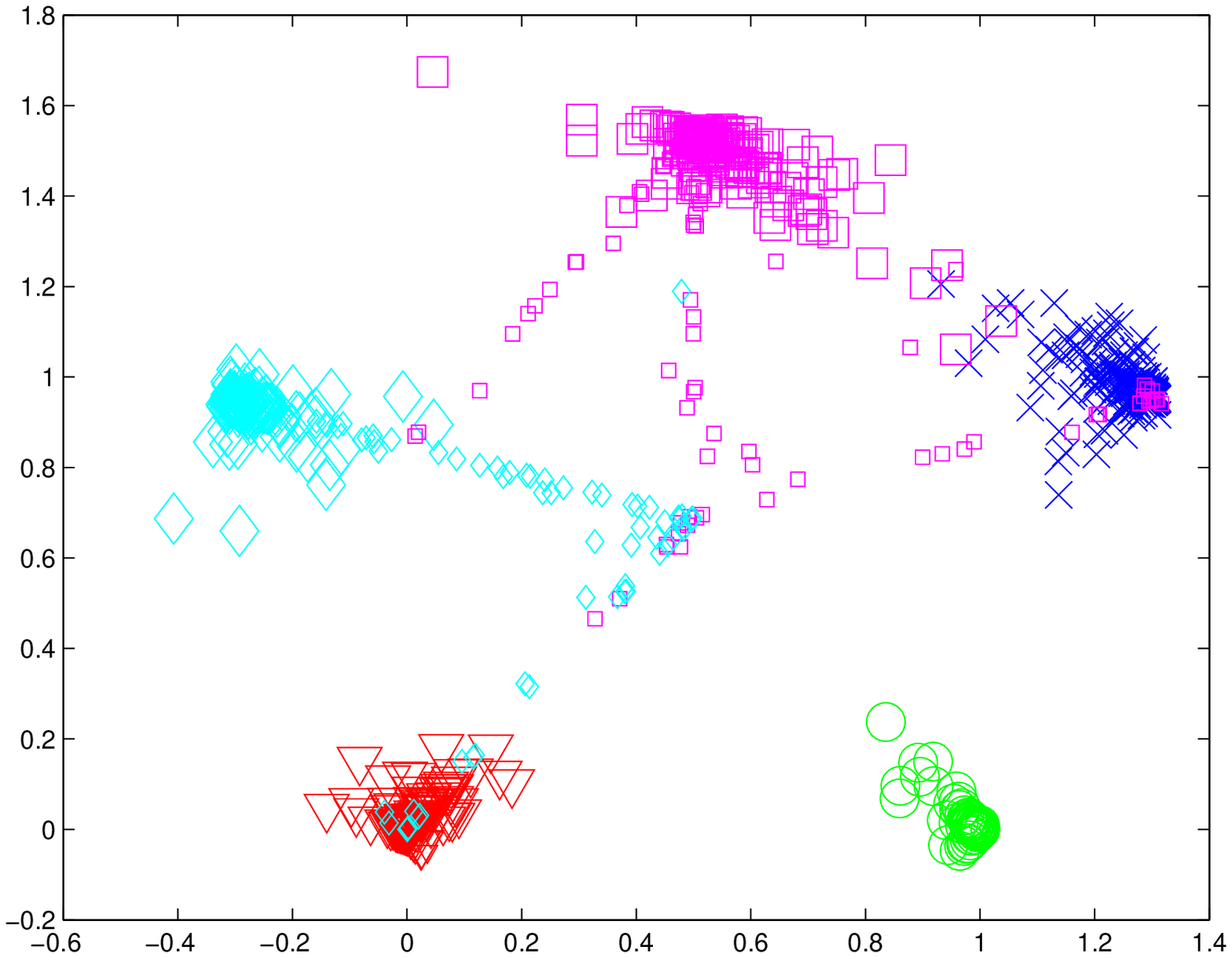,width=4.3cm}

\caption{Satimage data, first five classes, MLP with 30 hidden neurons and 0.05 regularization; right figure with additional 100 points for each class, generated by adding noise to selected vectors.}
\label{fig:satim}
\end{figure}

The Satimage data \cite{UCI} originally contained images of six types of soil from the Landsat satellite multi-spectral scanner. The 3x3 neighborhoods of a central pixels from 4 different spectra re provided as feature vector (36 dimensions). The last, mixed soil class, has been removed to make small figures more legible, leaving 5 classes only and 3397 training samples. An MLP with 30 hidden nodes and 0.05 regularization coefficient has been trained on this data, providing good separation of most data points (left plot, Fig. \ref{fig:satim}). Most errors are due to mixing of the class 3 and 4 vectors. How stable is this solution? One point from each class has been selected, and 100 noise points generated by placing a Gaussian with 3\% variance added, providing additional 500 points for display (right plot, Fig. \ref{fig:satim}). In some feature space areas reliability of classification is very high, with all 100 noise points staying within the cluster for triangles, circles and crosses. Many points generated near the vectors from the squares and diamonds class are in the region where none of the network outputs has strong value (center of Fig. \ref{fig:satim}). Other additional vectors are on the line between the corner representing wrong class, and the center, indicating that only one (wrong) output has value significantly greater than zero. Images of some vectors appear in the center of a wrong cluster, showing that the network is still too confident in its predictions, with sharp decision borders close to the data points. Recognizing the existence of such regions is obviously very important in safety critical applications.

\section{Discussion and conclusions}

Neural networks are used in various ways for data visualization. The activity of two hidden neurons of MLP or RBF networks may be displayed directly. Self-Organized-Maps and other competitive learning algorithms, neural Principal and Independent Component Analysis algorithms, autoassociative feedforward networks and Neuroscale algorithms are all aimed at using neural algorithms to reduce dimensionality of the data or to display it (for a summary of such visualization methods see \cite{Antoine}).
The visualization method presented here is rather different, since neural networks are not modified or used to display multidimensional data directly, but rather a projection method is introduced to elucidate the network function.
The method is applicable to any black box classification system that outputs some estimation of class memberships. Although linear projection cannot show all details of the higher dimensional data distribution (i.e. for more than 2 classes), it contains a lot of useful information. For two classes the images of data vectors appear in a square, with (1,0) and (0,1) corners coresponding to uniquely classified cases, (0,0) to unknown case (both outputs are close to zero), and (1,1) to cases in the overlapping regions. Such detailed information is unfortunately difficult to display in two dimensional plots for more than two-classes.

Images of the training data vectors mapped by MLP and RBF neural networks have been used here to show the dynamics of learning, to compare different network solution, inspecting the regions of the input space where potential problems may arise, to evaluate effects of regularization, to investigate stability of network classification under perturbation of original vectors and to place new data in relation to known data vectors, allowing for estimation of confidence that one may have in classification of a given vector. The best network solutions are not overconfident, but show large clusters of points around vertices of the polygon, without overlaps with clusters and with no vectors close to the center of the projection.

This type of visualization may also be combined with the
Receiver Operator Characteristic (ROC) curves that show detection rates for a given false alarm rate \cite{ROC}. Samples with images close to the polygon vertices correspond to the high probability assigned by the classifier. Leaving just those data vectors that are below the specified low detection rate will leave only images close to the polygon vertices. Moving to higher detection rates the number of errors observed is roughly inversely proportional to the slope of ROC curve. Scatterograms carry more information, showing what type of errors are made and allowing for quick identification of such data vectors.
The common practice of selecting the largest network output value as the class indicator leads to optimal decision borders only for well separated images in scatterograms; more accurate decision boundaries in the image space may be selected.

A number of other options remains to be investigated, including applications to visualization of dynamic data.
There is no reason why scatterogram images of the known data should not always be displayed as a part of the neural network output. Although such visualization may not open the black box completely, at least it adds some color to elucidate its function.

{\bf Acknowledgement.} I am grateful to dr N. Jankowski for discussions on network function visualization, in particular for the idea to use polygon corners for projections. Initial version of the Matlab software used for simulations presented in this paper has been developed by Mr M. Orlowski as a part of his MSc thesis.
This paper has been presented at  International Joint Conference on Neural Networks (IJCNN), 2003.



\vspace*{-10pt}
\begin{thebibliography}{99}

\bibitem{duchtnn}
W. Duch, R. Adamczak and K. Gr\c{a}bczewski,
{\em Methodology of extraction, optimization and application of crisp and fuzzy logical rules.}
IEEE Transactions on Neural Networks {\bf 12}: 277-306, 2001.

\bibitem{ROC}
J.A. Swets,
{\em Measuring the accuracy of diagnostic systems.}
Science {\bf 240}, 1285-93, 1988.

\bibitem{Bishop}
C. Bishop,
{\em Neural networks for pattern recognition}.
Oxford: Clarendon Press, 1994.

\bibitem{pdfversion}
PDF version of this paper and the Matlab files are available at:
http://www.phys.uni.torun.pl/kmk/publications.html

\bibitem{UCI}
C.L, Blake, C.J. Merz,
UCI Repository of machine learning databases,
http://www.ics.uci.edu/~mlearn/MLRepository.html.
University of California, Irvine, Dept. of Information and Computer Science, 1998-2003.

\bibitem{Netlab}
I. Nabnay and C. Bishop,
NETLAB software,
Aston University, Birmingham, UK, 1997.
http://www.ncrg.aston.ac.uk/netlab/

\bibitem{Antoine}
A. Naud, (1994):
{\em Neural and statistical methods for the visualization of multidimensional data}.
PhD thesis, Dept of Informatics, Nicolaus Copernicus University, 2001.
Available from http://www.phys.uni.torun.pl/kmk/publications.html


\end{thebibliography}
\end{document}